\documentclass{article}
\ifdefined\pdfminorversion
\fi
\usepackage{iclr2027_conference,times}
\usepackage[T1]{fontenc}
\usepackage[utf8]{inputenc}
\usepackage{amsmath,amssymb}
\usepackage{graphicx}
\usepackage{capt-of}
\usepackage{booktabs,tabularx,array,longtable}
\usepackage{xcolor,colortbl}
\usepackage{placeins}
\usepackage{needspace}
\usepackage{url}

\usepackage[pagebackref=true,breaklinks=true,colorlinks=true,bookmarks=false]{hyperref}
\usepackage{xspace}
\usepackage{wrapfig}
\usepackage{listings}
\usepackage{tcolorbox}

\usepackage{tocloft}
\usepackage[toc,page,header]{appendix}
\usepackage{adjustbox}
\usepackage{minitoc}

\usepackage[utf8]{inputenc} %
\usepackage[T1]{fontenc}    %
\usepackage{url}            %
\usepackage{booktabs}       %
\usepackage{amsfonts}       %
\usepackage{nicefrac}       %
\usepackage{microtype}      %

\usepackage{booktabs}       %
\usepackage{amsfonts}       %
\usepackage{nicefrac}       %
\usepackage{microtype}      %
\usepackage{mathtools}
\usepackage{multirow}
\usepackage{bm}
\usepackage{epsfig}
\usepackage{graphicx}
\usepackage{caption}
\usepackage{subcaption}

\usepackage{amsthm}
\usepackage{amsmath}
\usepackage{amssymb}
\usepackage{enumitem}
\usepackage{makecell}
\usepackage{wrapfig}
\usepackage{indentfirst}
\usepackage{verbatim}
\usepackage{color}
\usepackage{xcolor}
\usepackage{setspace}
\usepackage{array}
\usepackage{booktabs}
\usepackage{stackengine}
\usepackage{graphicx}
\usepackage{textgreek}
\usepackage{slashed}

\definecolor{deepred}{HTML}{940000}
\hypersetup{linkcolor=deepred}
\hypersetup{urlcolor  = [rgb]{0.4,0.15,0.95}}
\hypersetup{citecolor=[rgb]{0.4,0.15,0.95}}
\definecolor{linkbrown}{HTML}{B88155}
\usepackage{url}

\usepackage[T1]{fontenc}    %

\hypersetup{pdfauthor={},pdftitle={Can It Actually Work? Synthesizing Functional Testbeds for CAD Mechanisms}}

\iclrfinalcopy
\author{%
  \fontsize{9.5pt}{\baselineskip}\selectfont
  Wenqian Zhang\textsuperscript{1,$\dagger$}\quad
  Yangyi Huang\textsuperscript{2,$\dagger$}\quad
  Zigeng Xu\textsuperscript{1,$\dagger$}\quad
  Weiyang Liu\textsuperscript{2,$\ddagger$}\quad
  Zhen Liu\textsuperscript{1,$\ddagger$,*}\\[0.2em]
  \fontsize{9pt}{\baselineskip}\selectfont
  \textsuperscript{1}The Chinese University of Hong Kong (Shenzhen)\quad
  \textsuperscript{2}The Chinese University of Hong Kong\\[0.1em]
  \fontsize{9pt}{\baselineskip}\selectfont
  \textsuperscript{$\dagger$}Equal contribution\quad
  \textsuperscript{$\ddagger$}Equal advising\quad
  \textsuperscript{*}Corresponding author
}
\title{\vspace{-3.5mm}Can It Actually Work?\\Synthesizing Functional CAD Mechanisms}

\definecolor{draftcolor}{RGB}{84,84,100}
\newcommand{\draftnote}[2]{{\color{draftcolor}\textbf{[#1:} #2\textbf{]}}}

\newcolumntype{Y}{>{\raggedright\arraybackslash}X}

\newif\ifverifiedreferences
\verifiedreferencestrue

\usepackage{pifont}
\usepackage{tocloft}
\usepackage[toc,page,header]{appendix}
\usepackage{adjustbox}
\usepackage{minitoc}

\renewcommand \thepart{}
\renewcommand \partname{}

\newcommand{\benchname}{FunCADBench\xspace}
\newcommand{\harnessname}{FunCAD-Harness\xspace}

\begin{document}
\maketitle

\begin{center}
    \vspace{-2.4em}
    \href{https://thegenlab.ai/funcad/}{\color{linkbrown}\textbf{\texttt{thegenlab.ai/funcad/}}}
    \vspace{-0.6em}
\end{center}

\begin{figure}[!ht]
\centering
\includegraphics[width=\linewidth]{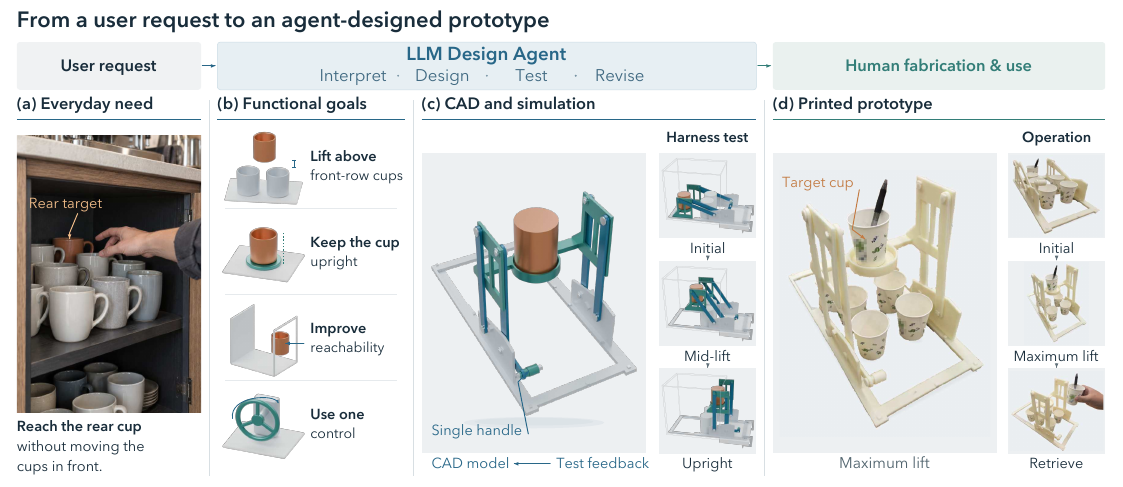}
\caption{\textbf{Functional CAD design from user request to physical prototype.} In the illustration, a user request for cup access is decomposed into functional goals that guide agent-based CAD construction and simulation-driven refinement. The resulting design is fabricated and operated by a human, with the printed prototype shown at maximum lift and during cup retrieval.}
\label{fig:teaser}
\vspace{1.5mm}
\end{figure}

\doparttoc 
\faketableofcontents
\begin{abstract}
\vspace{-0.5mm}
While LLM agents can generate visually plausible CAD assemblies, these designs often fail to fulfill their intended physical functions. We study \textit{functional CAD design}, a setting where agents interpret underspecified requests to construct and iteratively revise mechanisms that satisfy functional requirements in physical simulation. Verifying these requirements is nontrivial: different candidate mechanisms may realize the same function through entirely different components, controls, and operating sequences. To bridge this gap, we propose FunCAD-Harness, a framework that translates abstract requirements into candidate-specific physical tests and uses simulation feedback to guide design revisions. To systematically evaluate this capability, we introduce FunCADBench, a benchmark spanning 46 mechanism families with interdependent and multi-step operations. Experiments across frontier LLM agents show that FunCAD-Harness exposes subtle mechanical failures invisible to visual inspection, substantially improving functional success rates over standard simulation-augmented baselines. Ultimately, our work marks a step toward CAD agents capable of translating human intent into working mechanisms.
\end{abstract}

\section{Introduction}
\label{sec:intro}
\vspace{-1mm}

A longstanding ambition for generative AI is to turn human intentions into machines that can be built and made to work. Recent progress in LLM agents brings this ambition closer, enabling them to write modeling programs and operate CAD software to construct sophisticated mechanical designs for fabrication by 3D printing or CNC machining. Yet a visually plausible CAD model does not ensure that the resulting machine will fulfill the user's request, because its function depends on how its parts interact physically during operation. Even an assembly with valid parts and plausible articulation can fail to transmit a load, maintain contact, or return to a state that permits the next operation. A gripper, for example, may grasp an object successfully but leave it poorly positioned after release, preventing the next grasp. Because fabricating and assembling each candidate is costly, physical simulation offers a practical way to examine these interactions under specified inputs and loads and use the resulting feedback to revise the design.

Verifying a design through physical simulation, however, requires tests that reflect both the user's intended function and the way the candidate mechanism realizes it. Users often describe what they want a machine to accomplish without specifying all the behaviors needed to achieve it. Even when these requirements are made explicit, they do not uniquely determine an experiment, because the same function can be realized by mechanisms that operate differently. For an alarm that must trigger autonomously at a requested time, a gravity-powered design requires raising a weight, whereas a mainspring-powered design requires winding. A test designed around one mechanism can therefore be inappropriate for the other, even though the intended function is unchanged.

We study whether LLM agents can carry this process from an underspecified request to a CAD mechanism that satisfies its functional requirements in simulation, a problem we term \textit{functional CAD design}. To examine how far current agents can go on this task, we introduce FunCADBench, a benchmark spanning 46 mechanism families. Its tasks range from clamps and transmissions to mechanical counters, alarm clocks, and coin-sorting devices, whose functions depend on different combinations of geometry, force transmission, and changing contact states. Each task specifies a fixed request and an abstract functional test template while leaving the mechanical realization open. Success requires the resulting assembly to perform the complete requested behavior and satisfy the applicable geometric and physical constraints. Since repeated physical simulation can be costly for complex mechanisms, we develop a customized simulation environment to support efficient benchmark evaluation.

For this specific task, we develop FunCAD-Harness, which translates user requests into task requirements and abstract physical test templates. It then determines test conditions appropriate to the candidate mechanism and uses structured simulation feedback to guide design revision. Fig.~\ref{fig:teaser} illustrates this workflow through an open-ended cup-access design and its printed prototype.

With FunCAD-Harness, we evaluate frontier models and baseline workflows on FunCADBench and find that structured physical testing and diagnostic feedback improve task success. The model comparisons also show that strong visual 3D modeling capabilities do not necessarily translate into superior functional design. To better understand how agents approach this task, we analyze their tool use and observe that they rely primarily on CAD programs and simulation, with little use of rendered images. Testing the role of visual access directly, a separate experiment shows that substantial task success remains possible when it is removed from the constructor.

In summary, our contributions are:
\begin{itemize}[leftmargin=*,nosep]
\setlength\itemsep{0.35em}
    \item We formulate \textit{functional CAD design}, in which agents translate underspecified requests into mechanisms that fulfill their intended functions in simulation, and introduce FunCADBench, spanning 46 mechanism families, with a simulation environment for efficient physical evaluation.
    \item We develop FunCAD-Harness, which derives requirements and abstract test templates from user requests, constructs physical tests suited to candidate mechanisms, and uses structured simulation feedback to guide iterative design and revision.
    \item We systematically evaluate frontier agents and baseline workflows, demonstrating the benefits of physical testing and diagnostic feedback. Our analyses further reveal performance--cost tradeoffs and recurring challenges in functional mechanical design.
\end{itemize}

\vspace{-1mm}
\section{Related Work}
\label{sec:related}
\vspace{-1mm}

\textbf{Mechanical design and AI for engineering.}
Mechanical design connects intended functions to the structure and motion of mechanisms~\citep{reuleaux2013kinematics}. Computational approaches optimize mechanisms and structural topology~\citep{coros2013computational,bendsoe2004topology}, while learned models predict performance and generate candidate designs~\citep{umetani2018learning,regenwetter2022deep}. Co-design accounts for dependencies between morphology, motion, and control~\citep{ha2016task,yuan2022transform2act}, and between hardware and software under resource and compatibility constraints~\citep{carlone2019robot}. LLMs connect natural-language specifications to design generation, analysis, and optimization tools~\citep{makatura2023can}. Executable evaluation supports both benchmarking, as in EngDesign's cross-domain design tasks~\citep{guo2026toward}, and search, as in VLMgineer's simulation-guided evolutionary co-design of robot-mounted tool geometry and action waypoints~\citep{gao2026vlmgineer}. Closest to our setting, BesiegeField studies \emph{compositional machine design} from textual functional demands, assembling standardized components in a game-based simulator and using simulation rewards and state feedback to revise the assembly~\citep{zhang2026compositional}. Our CAD setting expands the design space beyond predefined blocks: agents jointly synthesize part geometry, assembly structure, and operating inputs to realize the requested functions, producing CAD models that provide a basis for eventual fabrication.

\textbf{Programmatic CAD and physically grounded assets.}
Programmatic CAD and graphics representations make modeling operations and part composition editable~\citep{ritchie2023neurosymbolic,wu2023cad,li2025cad}, with compilation, rendered views, and CAD-engine diagnostics providing feedback on geometric fidelity and modeling validity~\citep{alrashedy2025generating,gong2026toolcad}. Assembly benchmarks extend evaluation to functionality, manufacturability, and assemblability through executable geometry checks and drawing-based VLM judgments under supplied assembly graphs and engineering constraints~\citep{dong2026musebenchmarkingmanufacturablefunctional}. Articulated-asset generation further tests how parts interact, using agent-authored checks of contact, clearance, containment, and pose-dependent relations~\citep{zhou2026articraft}, or independent review and physical simulation of collisions, stability, and manipulation feasibility for coupled motions~\citep{cai2026physicallygroundedarticulated3d}. Whereas these approaches assess the realization and functional usability of specified objects, \benchname{} evaluates alternative mechanism architectures against a functional request's complete operating requirements, including load transmission and dependent or repeated operations.

\textbf{Agent workflows, harnesses, and executable evaluation.}
Agent interfaces and runtimes organize tool use, context, and execution feedback~\citep{yang2024sweagentagentcomputerinterfacesenable,wang2025openhandsopenplatformai}, while automated search optimizes workflows and full harness programs against supplied evaluators~\citep{zhang2025aflowautomatingagenticworkflow,lee2026metaharnessendtoendoptimizationmodel}. Executable checks also guide artifact construction within a workflow, from stage gates for robot-learning environments, interfaces, rewards, and training configurations~\citep{li2026harborharnessframeworkagentic} to geometric checks and finite-element stress and safety-factor feedback for structural CAD under prescribed loads, supports, and design-space constraints~\citep{berger2026physicsintheloophybridagenticarchitecture}. To accommodate generated implementations, software testing can adapt test inputs against a trusted reference and retain counterexamples for regression testing~\citep{kato2026gtddgenerativetestdrivendevelopment}, or bind fixed functional tests to implementation-specific identifiers~\citep{liu2025projecteval}. \harnessname{} extends candidate-specific test binding to operating mechanisms whose geometry, assembly, and inputs vary across designs. Registered requirements and abstract test templates remain fixed, while actuation, loads, and observations are bound to each candidate for simulation and source-geometry checks. This evaluates complete mechanism operation rather than structural response to a prescribed load case, and supplies feedback for structural revision.

\vspace{-1mm}
\section{FunCADBench: Benchmark for Functional CAD Design}
\label{sec:benchmark}

\vspace{-1mm}
\subsection{Functional CAD design}
\label{sec:task-specification}
\vspace{-1mm}

Mechanical design involves production cost, manufacturing processes, and regulatory requirements as well as mechanical behavior~\citep{textbook1,textbook2}. We focus on interpreting high-level user requests and determining whether a CAD mechanism fulfills the intended functions under appropriate physical conditions~\citep{zhang2026compositional}.
Formally speaking, each task of functional CAD design is specified by a natural-language \emph{request} describing a mechanism's intended use. The \emph{request} establishes its \emph{functional requirements}, which describe the behaviors to achieve, and \emph{fixed constraints}, which must be preserved throughout design. The agent produces a \emph{candidate design} together with \emph{declared physical assumptions}. Different mechanical implementations are acceptable as long as they satisfy these requirements; for example, a clamp may use different actuation mechanisms, but it must still close, hold an object, and release it.

\vspace{-1mm}
\subsection{Task Families and Taxonomy}
\label{sec:taxonomy}
\vspace{-1mm}

\begin{figure}[!t]
\begin{minipage}{\linewidth}
\vspace{-1.5mm}
\captionof{table}{Task taxonomy of \benchname.}
\label{tab:taxonomy}
\vspace{-1.75mm}
\centering
\footnotesize
\begin{tabularx}{\linewidth}{@{}>{\raggedright\arraybackslash}p{0.82in}YY@{}}
\toprule
Axis & Mechanical demands & Agent capabilities \\
\midrule
Motion primitives
& Motion conversion, transmission, phase, and repetition
& Spatial and physical reasoning about motion \\
\arrayrulecolor{black!20}\midrule[0.3pt]\arrayrulecolor{black}
Topology
& Connections, branches, and kinematic loops
& Reasoning about assembly constraints \\
\arrayrulecolor{black!20}\midrule[0.3pt]\arrayrulecolor{black}
Complexity
& Coupled subsystems, states, and operation sequences
& Task decomposition and revision planning \\
\arrayrulecolor{black!20}\midrule[0.3pt]\arrayrulecolor{black}
Contact
& Engagement, sliding, locking, and friction
& Reasoning about contact and friction \\
\arrayrulecolor{black!20}\midrule[0.3pt]\arrayrulecolor{black}
Geometric difficulty
& Feature construction, alignment, clearance, and assembly access
& CAD tool use and geometric verification \\
\bottomrule
\end{tabularx}
\end{minipage}
\par\medskip
\begin{minipage}{\linewidth}
\vspace{1mm}
\centering
\includegraphics[width=\linewidth]{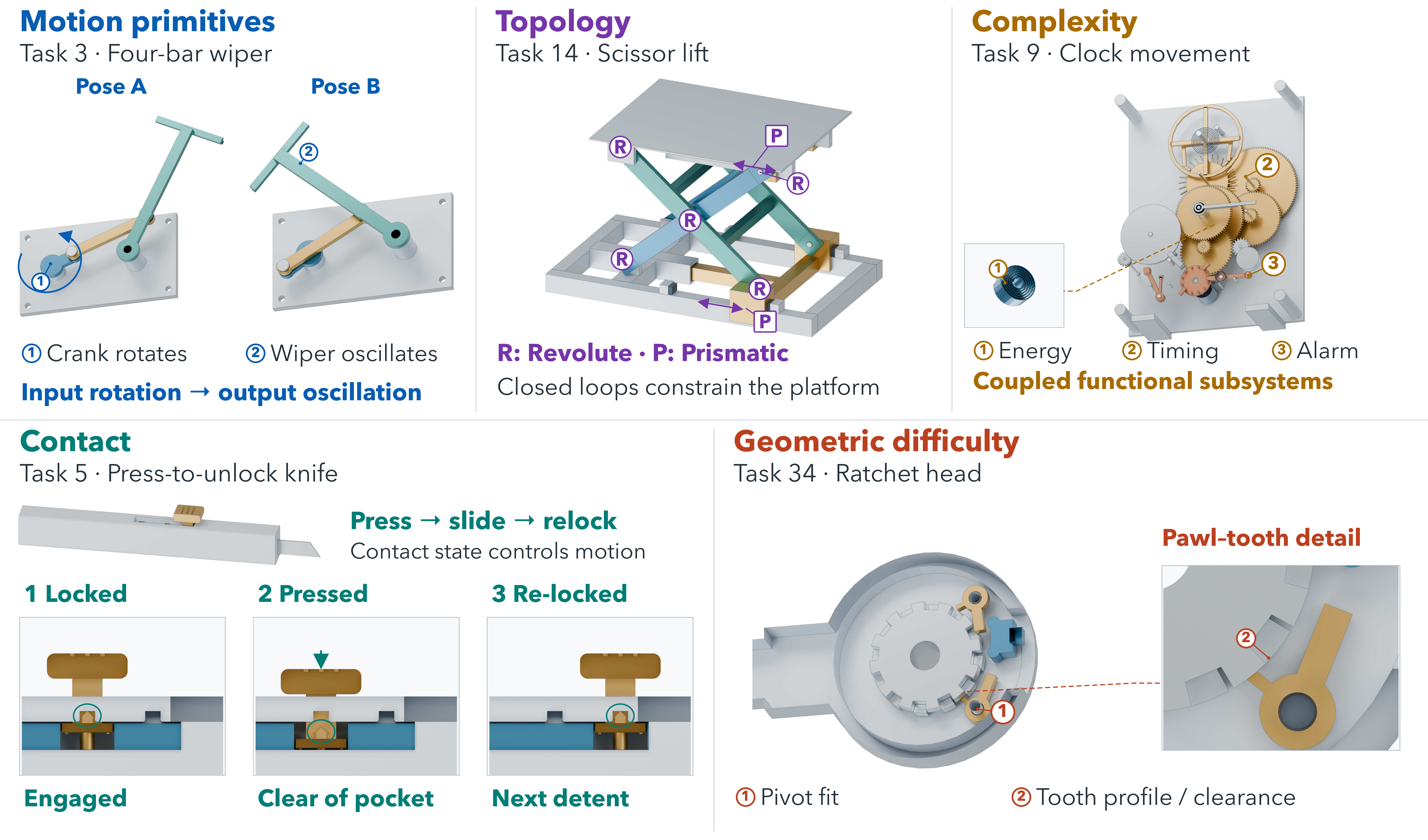}
\vspace{-7.5mm}
\caption{Taxonomy with representative mechanisms. Colors indicate functional parts and subsystems.}
\label{fig:representative-mechanisms}
\end{minipage}
\vspace{-3mm}
\end{figure}

\begin{figure}[t]
\centering
\includegraphics[width=\linewidth]{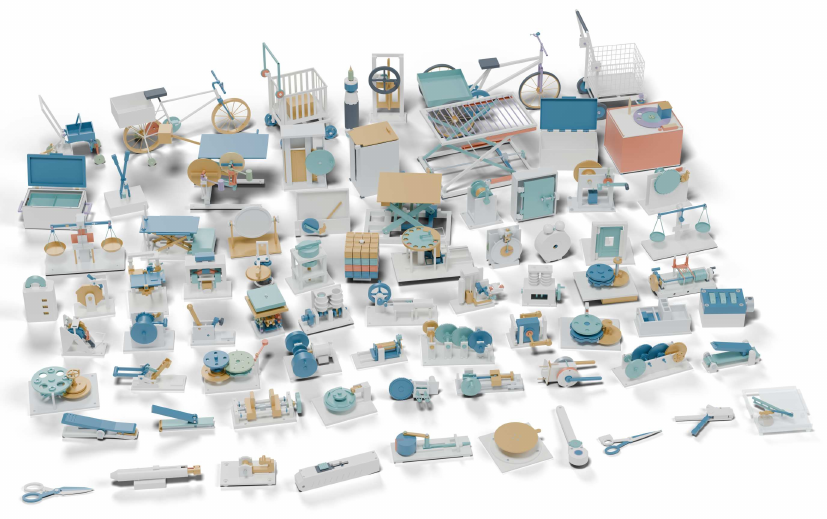}
\caption{\textbf{Mechanisms built by LLM agents in our experiments.} The 75 assemblies cover all 46 \benchname families and are rendered from their delivered CAD geometry (arranged for display, not to scale); 38 of them pass the full \harnessname evaluation.}
\label{fig:gallery}
\vspace{-2mm}
\end{figure}

To evaluate frontier LLM agents on this task across different mechanical demands, we introduce \benchname, a benchmark spanning 46 mechanism families (representative mechanisms are shown in Fig.~\ref{fig:representative-mechanisms}). These range from elementary assemblies to coupled transmissions and devices with sequential or repeated operation. Fig.~\ref{fig:gallery} shows mechanisms that LLM agents built for all 46 families in our experiments. Many of these families are drawn from familiar artifacts, such as clamps, whose intended uses can be described without specifying their internal mechanics. Several families pose open-ended mechanism design tasks that leave key architectural choices to the constructor. For example, the four-station indexing table (Task~40) must convert continuous input rotation into intermittent motion with dwell, without prescribing the indexing mechanism. We choose these families to cover different combinations of challenges along these axes--motion primitives, topology, complexity, contact, and geometric difficulty--each drawing on different agent capabilities (Table~\ref{tab:taxonomy}).

\vspace{-1mm}
\subsection{Physical Evaluation and Success Criteria}
\label{sec:physical_eval}
\vspace{-1mm}

Functional requirements specify what an assembly must accomplish, while its mechanical realization determines how those functions can be tested. Similar to the separation between test logic and implementation-specific parameters in ProjectEval \citep{liu2025projecteval}, we define abstract test templates for the required behaviors and bind them to each candidate mechanism, while keeping the requests and templates fixed across candidates for registered benchmark tasks. This binding identifies the candidate's operating interfaces (\textit{e.g.}, bicycle handlebars) and observation signals, and specifies the inputs, loads, physical attributes, and environmental conditions needed to execute the template. These choices reflect how the mechanism operates while preserving the task's required functions and fixed constraints. Parameters left open by the task may be selected or revised within its permitted scope (Appendix~\ref{app:permitted-adjustments}). Given these parameters, we run physical simulation and record feedback signals, including forces, positions, and contact or collision events, for the relevant mechanical modules. The assembly succeeds under the task's simulation protocol if it passes all required functional checks and all applicable source-geometry checks (\textit{e.g.}, collision checks) under its declared operating conditions.

\vspace{-1mm}
\subsection{Simulation Environment}
\vspace{-1mm}

Physical evaluation and repeated design revision require tests that can run within an agent's working budget. Our simulation environment uses Project Chrono~\citep{Chrono2016} for rigid-body dynamics and Bullet~\citep{coumans2015bullet} for collision detection. Since resolving detailed mechanical interactions can be costly, the environment also supports equivalent models. Gear couplings, for example, represent transmission between rotating components without resolving individual tooth contacts~\citep{coros2013computational,Chrono2016}.

\begin{wrapfigure}{r}{0.4\textwidth}
    \vspace{-4mm}
    \centering
    \includegraphics[width=0.9\linewidth]{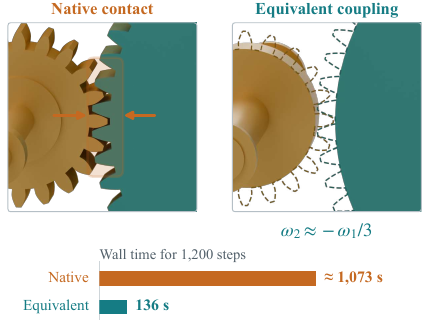}
    \captionsetup{font=footnotesize}
    \vspace{-.5mm}
    \caption{\footnotesize \textbf{Gear-transmission cost.}
    Native time is extrapolated from its recorded mean.}
    \label{fig:gear-runtime}
    \vspace{-5mm}
\end{wrapfigure}

When configuring a physical test, the agents choose between native simulation and a declared equivalent model for each interaction, recording its parameters and assumptions. If an equivalent model is used, it specifies force or torque laws for the corresponding source components, leaving the native solver to compute the resulting motion without prescribing output trajectories. Connections remain tied to finite regions of the source geometry and are released if those regions separate. The resulting motion provides the basis for functional checks, while separate audits examine the original CAD geometry at sampled poses. Appendix~\ref{app:physics} describes the interaction models, audit scope, and execution procedure.

\section{Agentic Harness for Functional CAD Design}
\label{sec:method}
\vspace{-1mm}

\begin{figure}[!t]
\centering
\includegraphics[width=\linewidth]{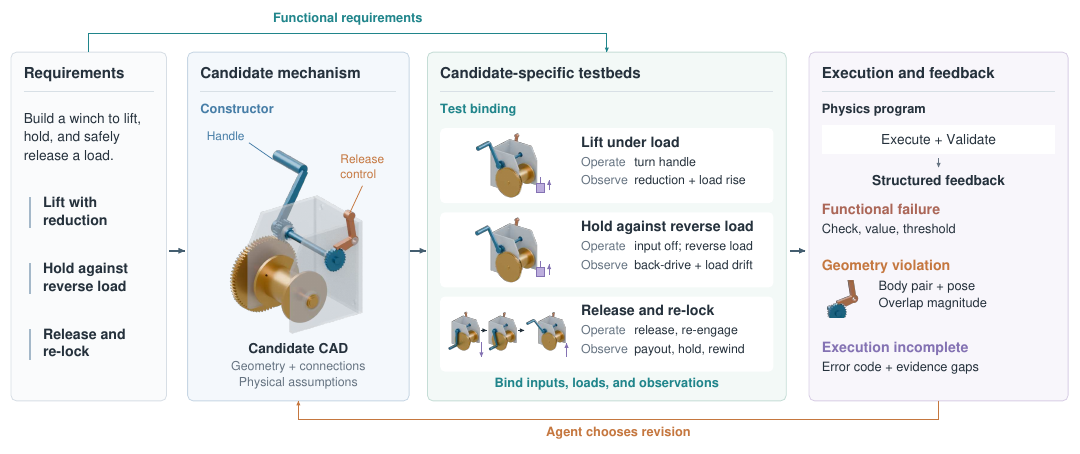}
\caption{Overview of \harnessname, illustrated with a releasable reduction winch (Task 42).}
\label{fig:overview}
\vspace{-5mm}
\end{figure}

A mechanism's design determines how its functions can be tested, and the test results in turn inform how the design should change. \harnessname organizes this interaction into a closed loop for developing and revising CAD mechanisms (Fig.~\ref{fig:overview}). Building on agentic CAD generation~\citep{zhou2026articraft} and simulation-based machine design~\citep{zhang2026compositional}, the harness pairs a constructor responsible for mechanism design with a test-binding agent responsible for preparing physical tests.

\vspace{-1mm}
\subsection{Functional requirements and test templates}
\label{sec:demand-parsing}
\vspace{-1mm}

The constructor starts from a text prompt specifying the functional requirements, which describe what the mechanism must accomplish while leaving its mechanical realization open. A request for an alarm clock, for example, may require it to trigger autonomously at a specified time without prescribing a particular escapement design. Each test template specifies the operations, observations, and success criteria for verifying the requested behavior, leaving mechanism-dependent details to candidate-specific binding. For the 46 registered FunCADBench tasks, both prompts and test templates remain fixed across candidate designs.

For requests outside the registered benchmark, the constructor first identifies the required functions, fixed constraints, and open operating parameters from the prompt, then proposes a test template. The test-binding agent checks its coverage and executability, and the constructor revises it accordingly. This additional template-authoring step is separate from the candidate-specific binding used to evaluate registered benchmark tasks.

\vspace{-1mm}
\subsection{CAD construction and physical testing}
\label{sec:method_construct_test}
\vspace{-1mm}

With the requirements and test templates in place, the constructor develops the CAD geometry, connections, operating inputs, and physical assumptions. The test-binding agent then binds each template to the candidate's operating inputs and observable states without changing its geometry or declared mechanics. For the winch in Fig.~\ref{fig:overview}, this involves identifying the handle, release control, and load attachment and configuring tests of lifting, holding, release, and re-locking. The parameters needed to execute these tests come from the task, geometry, design documentation, or task-authorized constructor choices. Their sources and any permitted adjustments are retained as described in Appendix~\ref{app:permitted-adjustments}.

Once the physical experiments have been specified through binding, the harness executes them without changing the CAD design between tests and checks the original geometry at sampled motion poses (Section~\ref{sec:physical_eval}). The execution program determines the outcomes from the recorded signals and applicable geometry checks.

\vspace{-1mm}
\subsection{Revision from physical feedback}
\label{sec:method_revision}
\vspace{-1mm}

These outcomes return to the constructor as per-function results, accompanied by physical signals, failed checks, geometry violations, and execution errors. Using this feedback, the constructor diagnoses failures and revises the geometry, connections, permitted operating parameters, or simulation representations. Since a revision may affect several functions, the test-binding agent updates the candidate-specific bindings for all required functions, and the harness reruns the full set of tests on the revised assembly.

\vspace{-1.5mm}
\section{Experiments and Results}
\label{sec:experiments}
\vspace{-1mm}

\subsection{Functional Design Performance}
\label{sec:model-comparison}

We evaluate GPT-6 Astra, Kimi K3, Fable 5.1, and Grok 4.6 with \harnessname{} on all 46 tasks (details in Table~\ref{tab:agent-configurations}), allowing one independent attempt per family within a three-hour active-time budget. Each model drives its own ReAct loop~\citep{yao2023react} for CAD construction, testing, and revision, with internal tests and revisions included in the generation process.

After each batch of generation runs, we evaluate the resulting assemblies in a separate batch using GPT-6 Astra with a one-million-token context allowance and medium reasoning. The final-evaluation model, context allowance, and reasoning setting remain the same across all generation models and comparison settings, including Grok + history. The evaluation agent configures candidate-specific physical tests under the task's required functions and fixed constraints, and the execution program determines their outcomes. Appendix~\ref{app:agent-configurations} provides the model settings and time-accounting details.

\begin{figure}[!htbp]
\centering
\includegraphics[width=\linewidth]{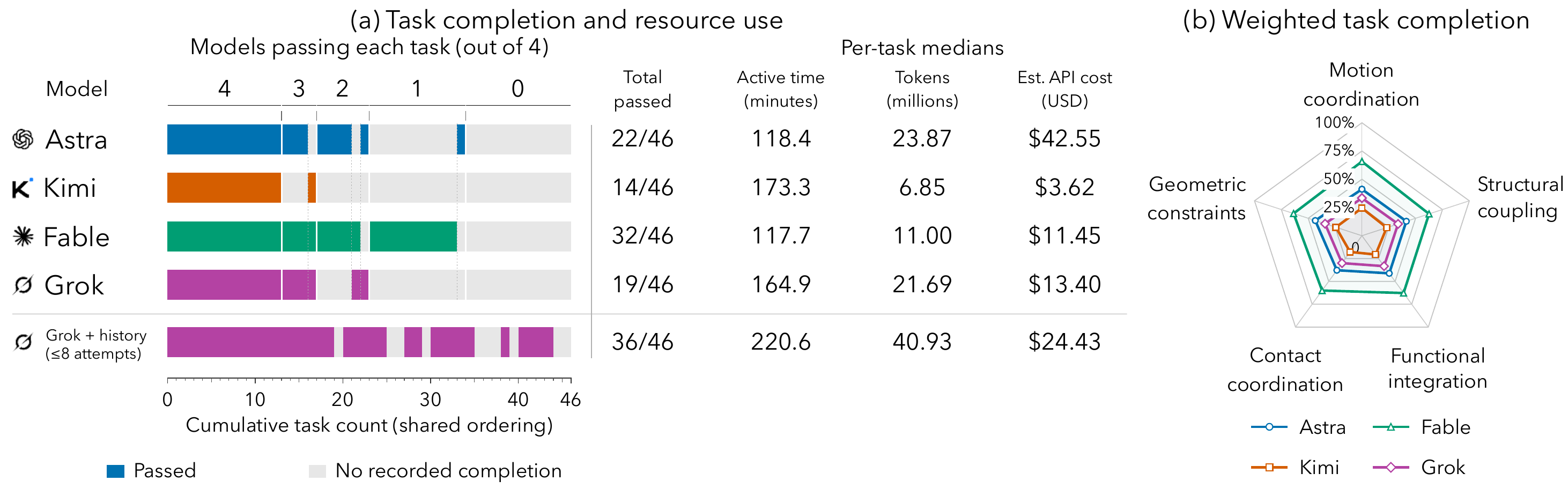}
\caption{Task success and mechanical demand coverage. (a) Tasks share an ordering across models. Colored segments indicate success, and gray segments indicate no recorded success. Individual outcomes appear in Table~\ref{tab:task-completion}. Time, tokens, and estimated constructor API cost are per-task medians. API cost covers recorded constructor usage and accumulates across attempts in the history setting (Appendix~\ref{app:outcome-aggregation}). (b) Success weighted by requirement grades 1--3 for each demand axis, normalized by the total weight over all 46 tasks (Appendix~\ref{app:demand-weighted-completion}). Task grouping and demand-weighted scores use the four single-attempt settings.}
\label{fig:model-comparison}
\vspace{-2.5mm}
\end{figure}

Across the four models, agents solve 34 of the 46 tasks (Fig.~\ref{fig:model-comparison}). All four succeed on most low-complexity tasks, but their performance diverges on more demanding mechanisms. Every task solved exclusively by Fable has medium or high complexity, although even this strongest model solves only five of the 15 high-complexity tasks. We also compare success across the benchmark's demand axes by weighting each task according to its requirement grade on the corresponding axis. All four models have their lowest weighted success rate on the contact axis, which covers engagement, release, friction, and changes in contact state.

We also evaluate Grok with up to eight attempts per task and access to records from earlier attempts. This setting solves 36 tasks, including six unsolved by any model in the default setting, at a lower median cumulative constructor API cost than Astra's default runs.

\subsection{Harness Comparisons}
\label{sec:harness-ablations}

To assess the contribution of structured physical testing, we compare \harnessname{} with three alternative workflows (Table~\ref{tab:harness-ablations}). In the \emph{requirements-only} and \emph{simulation-tools} baselines, agents choose their own tests and may implement their own simulations. The \emph{binary-feedback} variant retains the full harness workflow but returns only pass/fail simulation outcomes. We also test whether visual access contributes to construction and revision by repeating the Astra and Grok full-harness experiments with visual access removed only from the constructor. The other agents retain their standard visual access. All settings undergo independent final evaluation under the same protocol.

\begin{table}[t]
\centering
\small
\setlength{\tabcolsep}{5pt}
\caption{End-to-end completion on 46 functional CAD tasks under independent final evaluation. Success-set overlaps appear in Appendix~\ref{app:posthoc-evaluation}.}
\vspace{-1mm}
\label{tab:harness-ablations}
\begin{tabularx}{\linewidth}{@{}lYrr@{}}
\toprule
Setting & Provided support & Astra & Grok \\
\midrule
Requirements only & Requirements and agent-chosen verification & 1/46 & 3/46 \\
Simulation tools & Requirements, simulator, and tool guidance & 8/46 & 10/46 \\
Binary feedback & Full workflow with pass/fail simulation feedback & 0/46 & 8/46 \\
No constructor vision & Full harness without constructor visual access & 20/46 & 18/46 \\
Full harness & Tests of required functions and diagnostic feedback & \textbf{22/46} & \textbf{19/46} \\
\bottomrule
\end{tabularx}
\vspace{-1mm}
\end{table}

\textbf{Simulation access and whole-task success.}
Providing simulation tools improves success over requirements alone, and the full harness raises it further, from 8 to 22 tasks for Astra and from 10 to 19 for Grok. These gains span motion transmission and locking through mechanical contact. Both models additionally solve the rack-and-pinion gate, four-bar wiper, and hand-cranked whisk (Tasks~2--4), as well as the self-locking knife and spring-latch lock (Tasks~5 and~33).

\begin{figure}[!t]
\centering
\includegraphics[width=\linewidth]{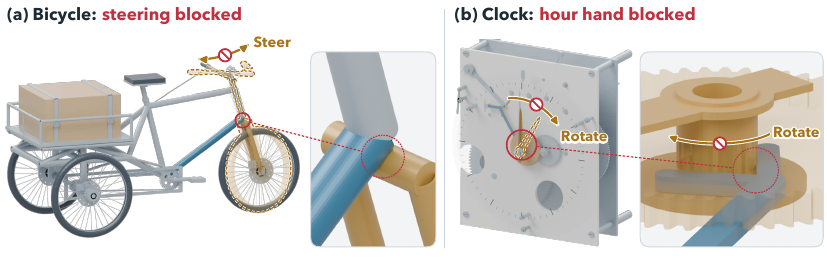}
\caption{Function-blocking interference in Astra designs produced without harness feedback. (a) The bicycle's steering assembly intersects the fixed frame (Task 7, \textit{requirements-only}). (b) The clock's hour wheel overlaps the alarm lever despite separate escapement and striker tests (Task 9, \textit{simulation-tools}). Red circles mark interference, and dashed silhouettes indicate intended poses. The clock inset uses schematic hand placement and sleeve length.}
\label{fig:astra-functional-failures}
\vspace{-3mm}
\end{figure}

Access to simulation does not, however, ensure that the tests cover the complete requested behavior. In the simulation-tools setting, Astra tests a clock's escapement for ten simulated minutes and its striker separately, but leaves the sequence of setting, triggering, and stopping the alarm while preserving timekeeping unverified. The delivered hour wheel also interferes with the alarm lever (Fig.~\ref{fig:astra-functional-failures}b).

\textbf{Retaining functional checks through revision.}
Functional checks can also be lost during repair. In the simulation-tools setting, Grok replaces a numerically invalid loaded test of a ratchet screwdriver (Task~13) with a short stepping test, leaving the drive and return-stroke requirements unchecked. A separate full-harness run retains these checks, revealing that all four stroke checks pass but both checks for backward slip of the output fail.

\textbf{Diagnostic feedback and workflow completion.}
When a test fails, diagnostic feedback helps the constructor identify what needs to change. The full harness reports failed requirements, measured discrepancies, and physical-validity violations that can distinguish failure causes and guide repairs. Removing these diagnostics reduces recorded success from 22 to zero tasks for Astra and from 19 to eight for Grok. Astra's binary-feedback runs include early delivery of partial results, so its zero successes include unfinished evaluation as well as failed checks. The benefit of diagnostic feedback extends to continued testing as well as design revision.

\vspace{-1mm}
\subsection{Design and Revision Behavior}
\label{sec:trajectory-analysis}
\vspace{-1mm}

\textbf{Repairing failures in repeated operation.}
Fable's gripper (Task~41) passes a short development test, but formal repeated-use testing reveals that the workpiece tips after release, causing the next grasp to miss it. The second grasp records 0\,N against a 10\,N minimum. Fable responds by widening the contact pads and support rest and reducing the permitted input-force limit from 50 to 35\,N. The revised design leaves the workpiece positioned for reuse, and its second grasp reaches 14.40\,N under the unchanged 5\,N external load. Both grasp--release cycles and the geometry audit pass (Fig.~\ref{fig:task41-design-trajectory}).

\begin{figure}[!t]
\centering
\includegraphics[width=\linewidth]{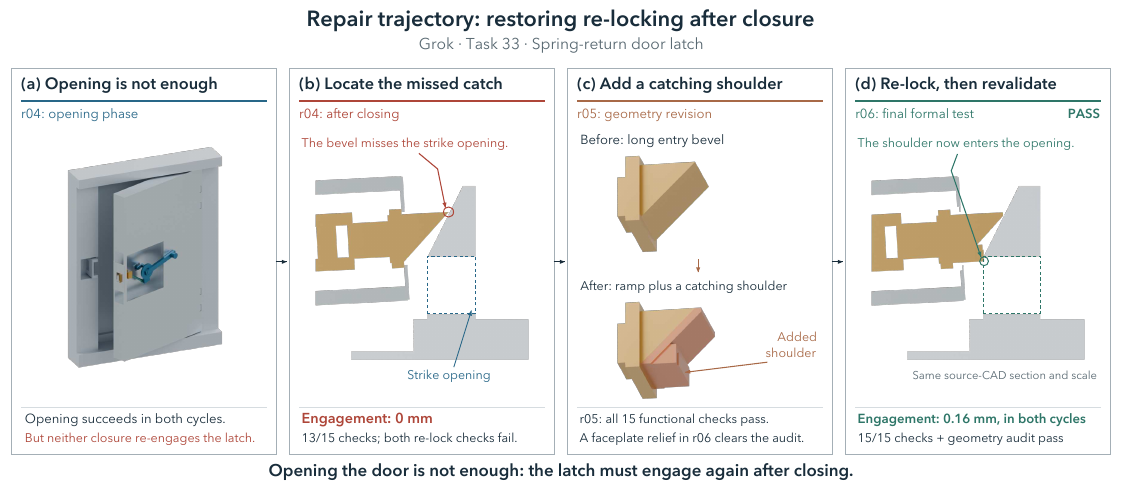}
\vspace{-3mm}
\caption{Grok repairs re-locking and local interference in a spring-return door latch (Task~33).}
\label{fig:task33-relocking-repair}
\vspace{-1mm}
\end{figure}

Restoring function can still leave geometric interference to resolve. In Grok's spring-return door latch (Task~33), the latch initially rests on the strike-plate edge after closure, failing both re-locking checks while passing the other 13 functional checks. Grok reshapes the entry bevel and adds a catching shoulder, restoring engagement after both closures and passing all 15 functional checks. It then adds local relief to the faceplate to resolve the remaining latch--faceplate interference. The final revision passes both the functional checks and the audit of the original CAD geometry under unchanged test parameters (Fig.~\ref{fig:task33-relocking-repair}).

\textbf{Verification strategies and visual access.}
We analyze tool use on 41 task families with recorded harness-tool execution times for all four models (Appendix~\ref{app:tool-use-analysis}, Fig.~\ref{fig:tool-profiles}). Fable and Grok devote larger shares of their calls to reading requirements, documentation, and source code. Astra devotes more to simulation, geometric checks, and analysis of test results and traces, with a higher share of simulation calls than Fable on 40 of the 41 matched tasks. Despite these differences in call distribution, simulation-related operations account for 75--85\% of recorded harness-tool execution time on average across tasks for every model.

Agents obtain geometric and physical information primarily through CAD programs and simulation, with little use of rendered images. Without visual access, Astra's constructor still succeeds on 20 of the 46 tasks, compared with 22 with visual access (Table~\ref{tab:harness-ablations}). Grok shows a similar pattern, completing 18 tasks without visual access compared with 19 with it. Substantial functional design success therefore remains possible without visual inspection during construction and revision. Appendix~\ref{app:no-vision} provides the protocol and task-level results.

\subsection{Tool Use and Execution Time}

\begin{figure}[!htbp]
\centering
\includegraphics[width=\linewidth]{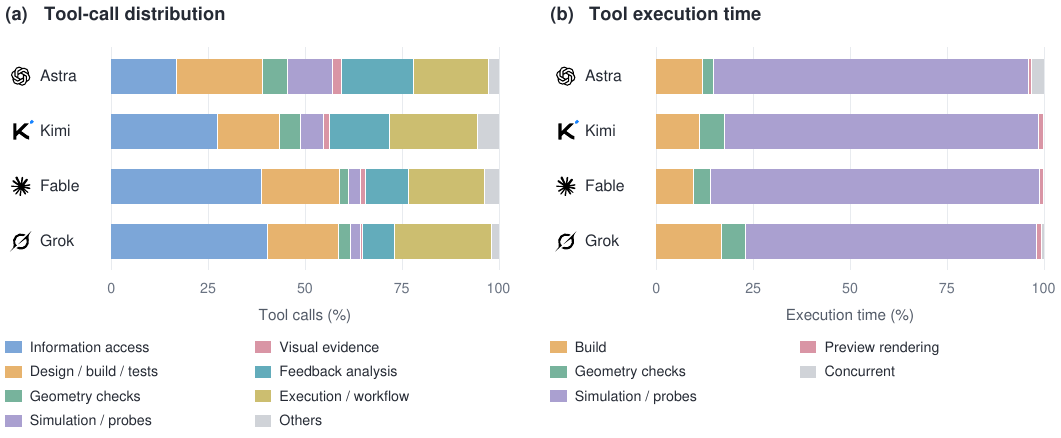}
\caption{\textbf{Tool use and execution time on 41 matched tasks.}
(a) Distribution of constructor tool calls.
(b) Distribution of recorded standard-tool execution time.
Simulation and cost probes are grouped together, and Concurrent denotes overlapping execution across categories.
Shares are computed within each task and averaged with equal task weights.
Call annotation, filtering, and timing details appear in Appendix~\ref{app:tool-use-analysis}.}
\label{fig:tool-profiles}
\end{figure}

We analyze tool use on 41 tasks with timing records available for all four models (Fig.~\ref{fig:tool-profiles}). Fable and Grok devote larger shares of their calls to reading requirements, documentation, and source code, whereas Astra devotes more to simulation, geometric checks, and feedback analysis. These differences in call frequency do not translate into equally different execution-time profiles. For every model, simulation and cost probes account for approximately 75--85\% of recorded tool execution time, even when they make up a small share of calls. Their cost highlights the importance of efficient simulation for repeated design revision. Across models, visual inspection remains infrequent, consistent with agents obtaining geometric and physical information primarily through CAD programs and numerical feedback.

\subsection{Real-world Fabrication}

\begin{figure}[!t]
\centering
\edef\prototypeheight{\the\dimexpr\linewidth*15/100\relax}

\begin{minipage}[t]{0.30\linewidth}
\centering
\includegraphics[height=\prototypeheight,trim=25bp 250bp 25bp 240bp,clip]{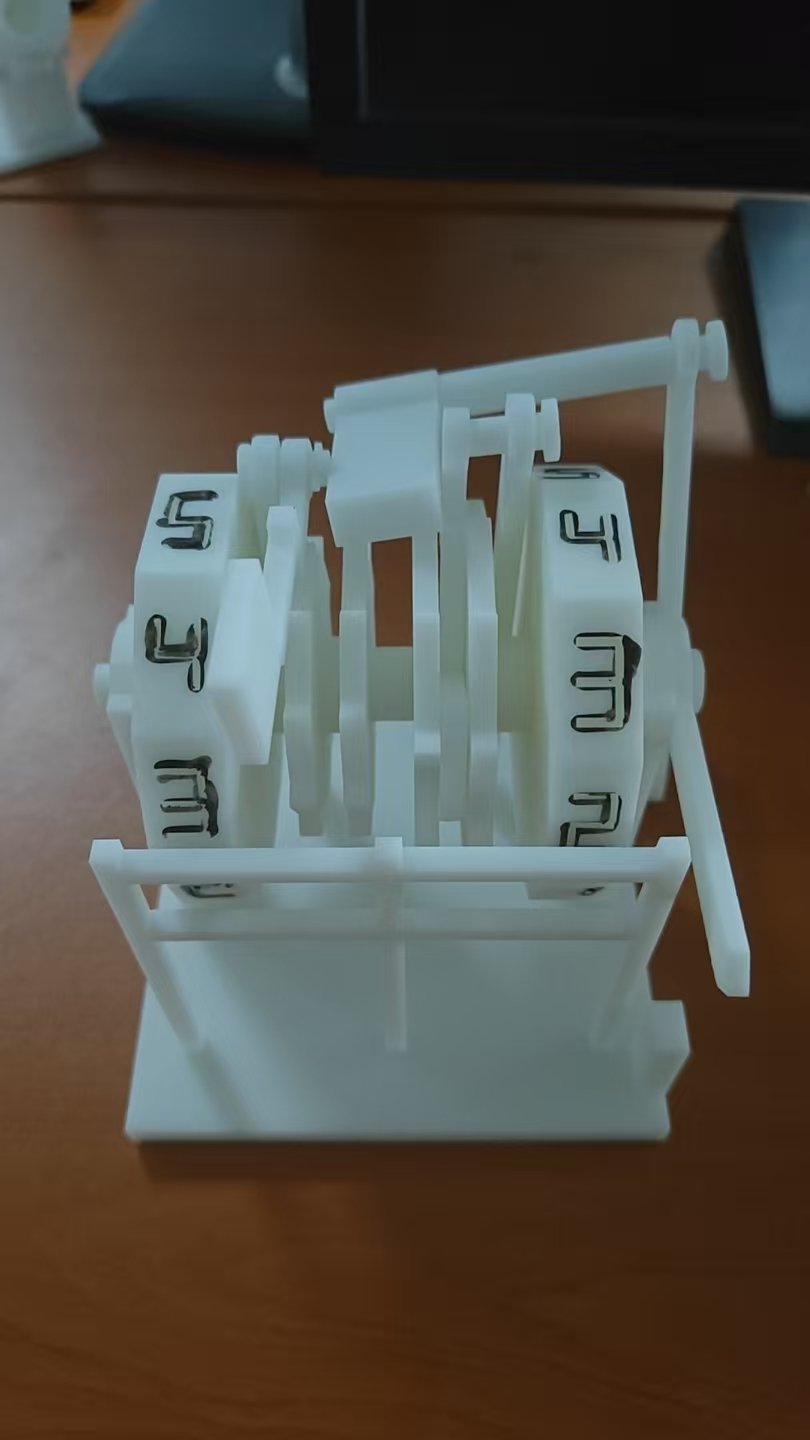}\hspace{2pt}%
\includegraphics[height=\prototypeheight,trim=25bp 240bp 25bp 250bp,clip]{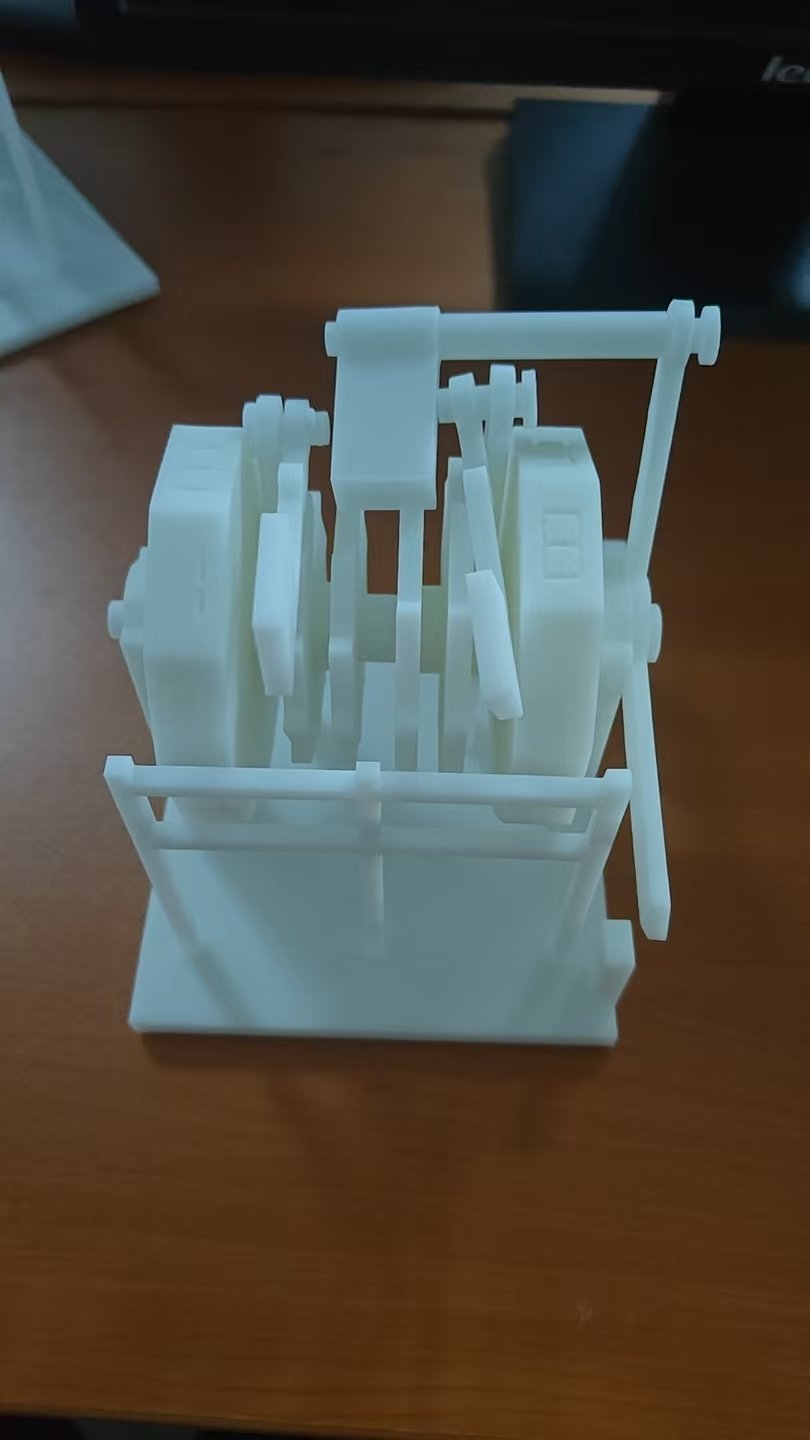}
\par\vspace{3pt}
{\small (a) Two-digit counter}
\end{minipage}\hfill
\begin{minipage}[t]{0.26\linewidth}
\centering
\includegraphics[height=\prototypeheight,trim=150bp 33.625bp 30bp 12.5bp,clip]{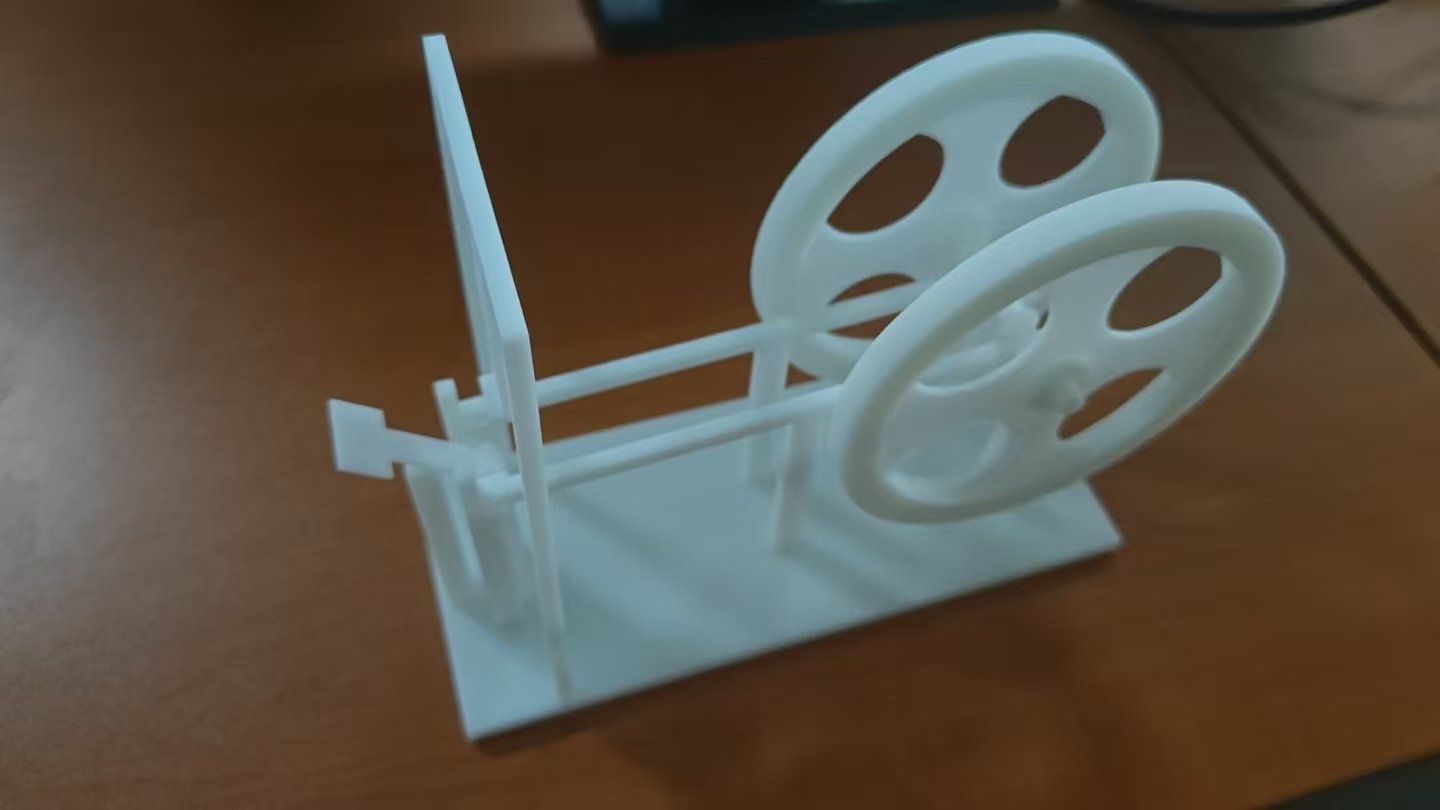}
\par\vspace{3pt}
{\small (b) Gravity timer}
\end{minipage}\hfill
\begin{minipage}[t]{0.42\linewidth}
\centering
\includegraphics[height=\prototypeheight,trim=30bp 697.5bp 30bp 165bp,clip]{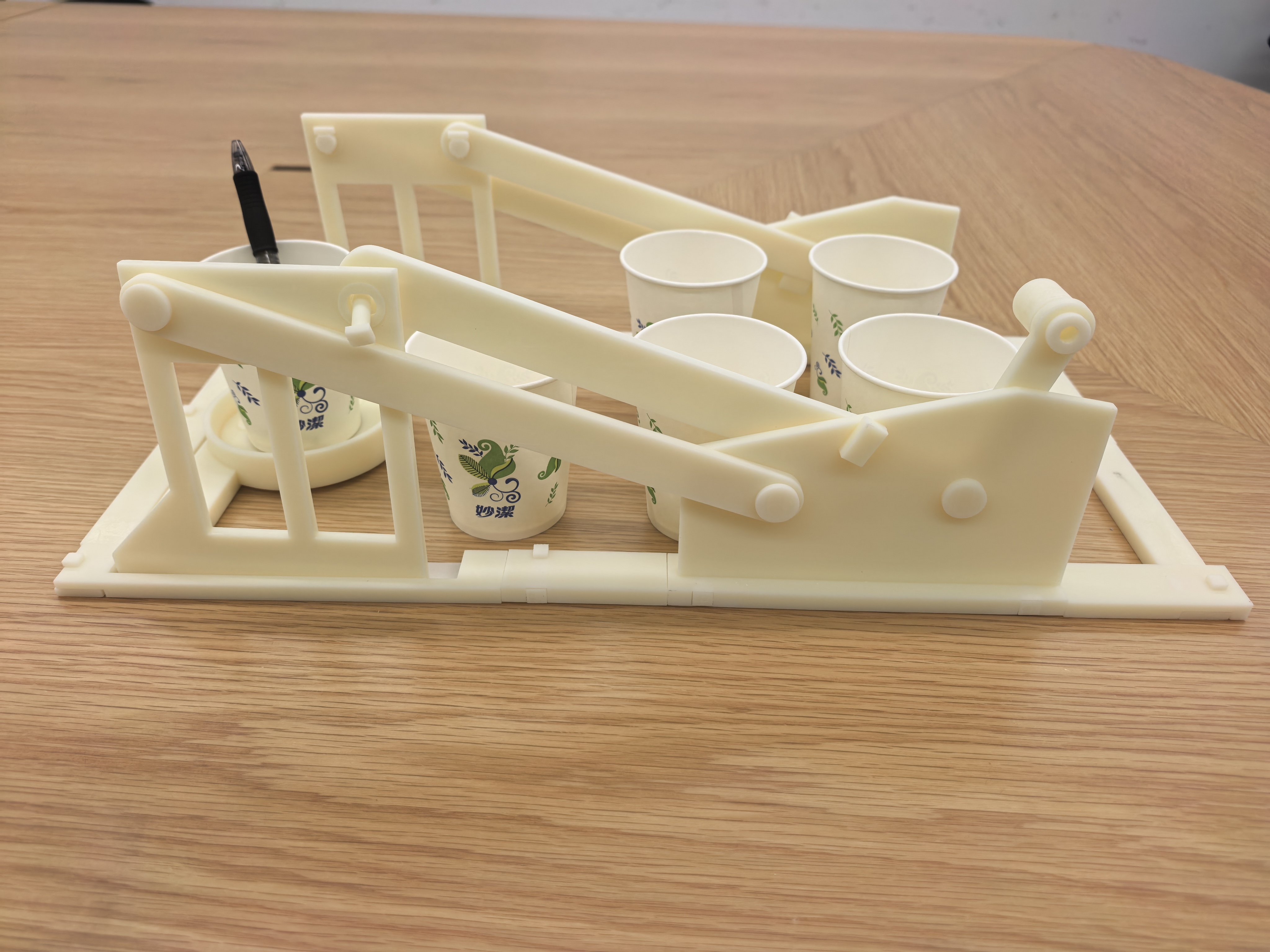}
\par\vspace{3pt}
{\small (c) Cup-access machine}
\end{minipage}

\caption{\textbf{Selected printed prototypes.}
(a) Front and rear views of the two-digit counter (Task~19).
(b) Gravity timer (Task~21).
(c) Initial configuration of the open-ended cup-access machine.
The pen marks the target cup at the rear.
Figure~\ref{fig:teaser} shows the raised configuration and retrieval.
Task numbers follow Appendix~\ref{app:mechanism-catalog}.}
\label{fig:printed-prototypes-cup}
\end{figure}

We 3D-printed and assembled prototypes to complement the simulation-based evaluation with qualitative examples of fabrication and assembly. Figure~\ref{fig:printed-prototypes-cup} shows selected prototypes, and supplementary videos document their operation. Additional examples appear in Figure~\ref{fig:printed-mechanisms} in Appendix~\ref{app:printed-mechanisms}.

\vspace{-1.5mm}
\section{Discussions}
\label{sec:discussions}
\vspace{-1.5mm}

\textbf{From simulation to fabrication.}
Turning a simulated mechanism into a working machine requires accounting for uncertainty in both physics models and fabrication. Contact and friction approximations, manufacturing tolerances, assembly misalignment, and material variation can change whether parts slide, engage, or jam. Experiments with selected prototypes could calibrate uncertain parameters and identify missing effects, following approaches that calibrate simulators from real-world robot trajectories~\citep{chebotar2019simopt}. Testing over plausible parameter ranges would then assess robustness to fabrication and operating conditions. The same prototype measurements could also help adapt world models, which offer fast predictions of rigid, articulated, and deformable interactions~\citep{huang2026pointworld,yang2025physworld}. Such adaptation could capture effects that are difficult to model explicitly and make repeated design tests more affordable.

\textbf{Coverage of functional tests.}
Improving simulation fidelity does not by itself ensure that the tests cover the intended behavior. Functional CAD design shares this challenge with project-level software generation in ProjectEval~\citep{liu2025projecteval}, where a high-level request must be translated into an artifact whose components work together. In both settings, ambiguous requirements and incomplete test suites can leave functions unchecked, while component tests may miss interactions. For mechanical systems, these interactions also depend on loads, initial states, and operation sequences, which determine contacts and the states available to later operations. Future harnesses could cover more of these conditions by combining requirement review and reusable engineering tests with automated searches for failure cases within the intended operating range~\citep{formica2024hecate}. The resulting tests could guide revisions and check that later designs preserve previously verified behavior.

\textbf{Multimodal interface fluency and physical reasoning.}
Despite its impressive visual 3D reconstruction capabilities~\citep{sun2026harmony}, Astra does not dominate functional CAD design. Fable solves more tasks, while Grok with additional attempts and retained history exceeds the best default result at a lower median cumulative constructor API cost than Astra. This contrast raises the possibility that visual modeling performance partly reflects multimodal interface fluency rather than a corresponding advantage in physical reasoning. Our agents obtain geometric and physical information primarily through CAD programs and simulation, and Astra retains substantial task success when its constructor has no visual access. Given that targeted post-training can develop proficiency in computer-use and CAD workflows~\citep{ramrakhya2025autoplay,gong2026toolcad}, training on these design--test--repair trajectories could improve functional design capabilities. Harnesses that help agents reuse prior designs and feedback could further improve the cost--performance tradeoff.

\vspace{-1.5mm}
\section{Concluding Remarks}
\label{sec:conclusion}
\vspace{-1.5mm}
Turning a high-level request into a functional CAD mechanism requires connecting the intended behavior to physical tests that reflect how the design operates. Through \benchname{} and its 46 mechanism families, we study this connection and develop \harnessname{} to use the resulting feedback for design revision. Our experiments show that organizing physical testing around the requested functions substantially improves whole-task success over providing simulation tools alone, helping agents repair failures in coupled and repeated operations. These findings highlight the importance of integrating functional verification into generation as LLM agents move from constructing 3D geometry toward designing mechanisms for practical use.

\ifverifiedreferences
  \bibliographystyle{iclr2027_conference}
  \bibliography{iclr2027_conference}
\else
  \section*{References}
  \draftnote{CITE}{To be completed with verified primary-source references. The reading leads in the draft have not yet been verified.}
\fi

\clearpage
\appendix

\addcontentsline{toc}{section}{Appendix} %
\renewcommand \thepart{} %
\renewcommand \partname{}
\part{\Large{\centerline{Appendix}}}
\parttoc

\newpage
\section{Prompts and Model-Visible Materials}
\label{app:prompts}
\subsection{Task prompts and constructor instructions}

Each of the 46 registered task prompts combines family-specific
functional requirements with shared CAD-delivery instructions.
The requirements specify the intended use, required functions, permitted
operations, and fixed constraints, while leaving the choice of geometry
and assembly structure to the constructor.
Attempts on the same registered family use the same task text;
additional harness instructions describe the available tools and provide
run-specific information, including the execution budget.
Experiment-specific task adjustment (operating/performance parameters) are recorded separately from the
original registered text.
Fig.~\ref{fig:overview} presents an abbreviated description of the Task 42
winch task, whose functional requirements include lifting with reduction,
holding against reverse load, and release followed by re-locking.

The box below summarizes the principal instructions provided to the
constructor during the experiments, retaining their design, delivery,
and verification requirements while omitting file organization and
execution details.

\begin{tcolorbox}[
    colback=gray!5,
    colframe=gray!60,
    title=\textbf{Constructor instructions (summary)}
]
Construct a complete and recognizable CAD assembly that satisfies the
required functions and preserves the task's fixed constraints.
Include all moving parts, supports, connections, and internal physical
effects required for its intended functions.

Provide rebuildable modeling source, a complete STEP assembly, the
information needed for physical simulation, and concise operating
instructions.

Identify the actual controls and their operating directions, the normal
operating sequence, observable physical states, and the units and basis
of relevant parameters.

Use the modeling representations permitted by the workflow, and document
the sources of their parameters and the limits of their applicability.

Support functional claims with execution evidence; do not substitute
prescribed output motion, fabricated measurements, or undeclared
fixation for the mechanism's functionality.

Leave formal test binding to the separate test-binding agent, who maps
the test requirements to the actual design's components, operations,
measurements, and parameters.

Use execution feedback to revise the design, and submit the updated
candidate for renewed binding and evaluation.

Do not assign a formal pass verdict yourself; the harness determines
completion from execution evidence.
\end{tcolorbox}

These instructions separate mechanical construction from formal test
binding: the constructor delivers the mechanism and its operating
instructions, the test-binding agent binds the formal tests, and the
program determines their outcomes from execution evidence.

\subsection{Harness Tool Descriptions and Returned Information}
\label{app:tool-materials}

The harness provides tools for building, previewing, checking, and
simulating candidate mechanisms.
Formal evaluation combines source preparation, independent test binding,
physical execution, and source-geometry auditing.

Source preparation loads the delivered geometry and physical scene for
evaluation.
After independent binding, interface checks verify that the required
inputs and observations are connected to the prepared model; passing
these checks does not establish functional success.
Physical execution produces the signals used to evaluate functional
criteria, complemented by audits of the original geometry at the initial
configuration and sampled motion poses.

The following excerpt shows the structured result of the second formal
evaluation of the Task 3 wiper.
The excerpt retains all five functional checks and the geometry-audit
outcome, while omitting assertion expressions and ancillary metadata.

\begin{tcolorbox}[
    colback=gray!5,
    colframe=gray!60,
    title=\textbf{Example structured evaluation result: Task 3}
]
\footnotesize
\begin{verbatim}
{
  "status": "PASS",
  "revision": 2,
  "tests": [{
    "test_id": "continuous_wiping",
    "execution_status": "completed",
    "physical_pass": true,
    "assertions": [
      {"id": "complete_sweeps", "passed": true,
       "observed_operands": [4.0, 3.0]},
      {"id": "links_attached", "passed": true,
       "observed_operands": [1.4331800428598694e-06, 0.0002]},
      {"id": "actual_input_cycles", "passed": true,
       "observed_operands": [4.0, 3.0]},
      {"id": "same_phase_return", "passed": true,
       "observed_operands":
         [3.468221060552712e-05, 0.017453292519943295]},
      {"id": "total_input_progress", "passed": true,
       "observed_operands":
         [28.192713231822275, 18.84955592153876]}
    ],
    "geometry_audit": {"passed": true}
  }],
  "complete_instance_pass": true
}
\end{verbatim}
\end{tcolorbox}

In each \texttt{observed\_operands} pair, the first value is the measured
statistic and the second is its comparison bound.
Cycle counts and accumulated rotation must meet or exceed their lower
bounds, while joint separation and return error must not exceed their
upper bounds.
Lengths and angles are reported in metres and radians, respectively.
All five functional checks and the source-geometry audit passed, and the
harness marked the candidate as complete under the tested conditions.

\subsection{Test-Binding Instructions}

The final evaluation agent instantiates the task template for the submitted
mechanism by identifying its operating inputs, loads, and observable
physical states.
The constructor's operating instructions inform this process, but the
correspondence is checked against the candidate's actual geometry and
mechanics.
For example, testing a toolbox requires identifying the moving lid, its
hinge, and a measurement of its opening angle; component names alone do
not establish these relationships.

\begin{tcolorbox}[
    colback=gray!5,
    colframe=gray!60,
    title=\textbf{Test-binding instructions (summary)}
]
Inspect the delivered candidate and map each required function to the
components, operating inputs, and measurements needed to test it.

Preserve the delivered geometry and declared mechanics when constructing
the test binding.

Specify the driven connections, loading conditions, effort schedules,
and observation signals according to the candidate's normal operation.

Record each test parameter together with its value and source, including
task requirements, geometric measurements, source documentation, or a
declared constructor decision.

Apply permitted parameter adjustments while retaining the original
values, revised values, and reasons for the changes.

After each design revision, ensure that the binding and evaluation
evidence correspond to the updated candidate.
Do not combine checks from different candidate configurations to claim
completion of a single design.
\end{tcolorbox}

The test-binding agent configures the candidate-specific experiment, while the
program determines its outcome from execution evidence.

\paragraph{Example of candidate-specific binding.}
The following excerpt shows the final evaluation agent's binding for the
second revision of the Task 3 wiper.
Comments marked FILLED identify entries supplied when constructing the executable binding, including numerical parameters and component references. The provenance of these entries is recorded separately.
Loading code and additional observation routines are omitted.

\begin{tcolorbox}[
    colback=gray!5,
    colframe=gray!60,
    title=\textbf{Candidate-specific binding example: Task 3}
]
\footnotesize
\begin{verbatim}
PROTOCOL = {
    "period_s": 1.8333333333333333,          # FILLED
    "repeats": 3,                          # FILLED
    "input_speed_rad_s": 6.283185307179586, # FILLED
    "input_torque_nm": 2.0,                # FILLED
    "gain": 0.02,                          # FILLED
    "sweep_rad": 1.53588974175501,          # FILLED
    "return_tolerance_rad":
        0.017453292519943295,              # FILLED
    "min_input_turn_rad":
        6.283185307179586,                 # FILLED
    "max_joint_separation_m": 0.0002,       # FILLED
}

frame = bodies["frame"]  # FILLED: reference body

model.actuators["input"] = Actuator(
    bodies["crank"],     # FILLED: driven body
    "body_torque",
    reference=frame,
    axis=(0.0, 0.0, 1.0),  # FILLED: drive axis
    react_on_reference=False,
)

model.signals["input.angle_rad"] = Signal(
    bodies["crank"],     # FILLED: observed input body
    "body_angle",
    reference=frame,
)

model.signals["wiper.angle_rad"] = Signal(
    bodies["rocker"],    # FILLED: observed output body
    "body_angle",
    reference=frame,
)
\end{verbatim}
\end{tcolorbox}

The operation and observation names, such as \texttt{input} and
\texttt{wiper.angle\_rad}, belong to the fixed template interface.
The test-binding agent connects these interfaces to the candidate's actual crank,
rocker, and reference frame.
The numerical protocol implements the constructor's recorded parameter
request, with the provenance of each value retained alongside the binding.
Here, \texttt{period\_s} denotes the observation allowance per required
cycle, not the commanded input-shaft period.
With three required cycles, the total observation duration is 5.5 seconds.

\subsection{Permitted Operating and Performance Adjustments}
\label{app:permitted-adjustments}

The task distinguishes required functions and explicit constraints from operating or performance parameters that the constructor may adjust. For example, a clamp task may permit changes to applied load or operating duration while still requiring closing, holding, and release. The constructor selects these values within the task’s permitted scope. The test-binding agent incorporates them into the test protocol and records their original values, revised values, and the reason for each change; programmatic validity checks remain in force. Completion is assessed under the final tested conditions, which are reported with the adjustment history.

\paragraph{Example of a constructor-requested adjustment.}
After the first formal Task 3 evaluation failed its motion checks, the
constructor requested an adjustment to the operating protocol.
The following example combines the recorded parameter request with its
accompanying decision rationale.
Only the changed parameters are shown, and the rationale is condensed
for presentation.

\begin{tcolorbox}[
    colback=gray!5,
    colframe=gray!60,
    title=\textbf{Constructor-requested adjustment: Task 3}
]
\textbf{Recorded action:} \texttt{adjust\_task}.

\medskip
\textbf{Requested parameter values:}
{\footnotesize
\begin{verbatim}
{
  "parameters": {
    "period_s": 1.8333333333333333,
    "gain": 0.02
  }
}
\end{verbatim}
}

\textbf{Preserved targets.}
Retain the commanded input speed of 60 rpm, the finite torque cap
of \(2\,\mathrm{N\,m}\), at least three complete cycles, an
\(88^\circ\) sweep, a maximum same-phase return error of \(1^\circ\),
and a maximum connection separation of \(0.2\,\mathrm{mm}\).

\medskip
\textbf{Changes from the previous protocol.}
Reduce the speed-control gain from
\(0.8\) to \(0.02\,\mathrm{N\,m\,s/rad}\).
Increase the observation allowance per required cycle from
\(1.5\) to approximately \(1.8333\) seconds, extending the total
observation duration from \(4.5\) to \(5.5\) seconds.

\medskip
\textbf{Worker rationale, condensed.}
The constructor attributed the repeated input reversals in the
preceding formal trace to excessive speed-control gain.
It requested gentler speed correction and a longer observation window.

\medskip
\textbf{Recorded trade-off.}
Accept slower speed correction, greater load-dependent speed error,
and a longer observation window while preserving the functional
targets and source geometry.
\end{tcolorbox}

The final evaluation agent implemented both requested changes in the
updated binding.
The remaining protocol parameters and source CAD geometry were unchanged.
The second formal evaluation passed all five functional checks and the
source-geometry audit, as illustrated in
Appendix~\ref{app:tool-materials}.
This example documents an operating-parameter adjustment and its
subsequent evaluation; it does not isolate the effect of either
parameter change individually.

\subsection{Template Authoring and Review for Open-Ended Requests}

Registered tasks reuse fixed functional templates, whereas open-ended
requests require templates to be constructed for the requested functions.
The constructor proposes the template, and an final evaluation agent
checks it before candidate-specific binding.

\begin{tcolorbox}[
    colback=gray!5,
    colframe=gray!60,
    title=\textbf{Template-authoring instructions (summary)}
]
Decompose the request into distinct functional requirements, each with
an observable outcome and an explicit pass criterion.

For each requirement, specify the operating conditions, input sequence,
measurements, and parameters to be determined for the candidate mechanism.

Preserve dependencies between operations: when a later function relies
on an earlier operation, evaluate it from the physical state produced
by that operation.

Define requirements in terms of the behavior the mechanism must achieve,
rather than the internal steps of a particular implementation.
For a mechanical alarm clock, the alarm triggering at the set time is
a functional requirement, whereas the escapement engaging a particular
tooth describes an implementation detail.

Reuse the shared execution components and submit the proposed template
for independent review.
\end{tcolorbox}

\begin{tcolorbox}[
    colback=gray!5,
    colframe=gray!60,
    title=\textbf{Template-review instructions (summary)}
]
Check that the template covers the requested functions and preserves
the task's fixed constraints.

Verify that the proposed operations, observations, and pass criteria
express the intended behavior and can be implemented through the
available interfaces.

Return missing or incorrect test definitions to the constructor
for revision.

If a required function cannot be evaluated using the available operations
or observations, record the limitation and leave that function unverified.
\end{tcolorbox}

The constructor may revise the template, but template review and formal
instance binding remain independent responsibilities.
A candidate is not considered complete while any required function
remains unverified.

\subsection{Materials for the Comparison Settings}

The four settings in Table~\ref{tab:comparison-condition-materials} vary the support provided for simulation, functional evaluation, and diagnosis. Requirements only does not prohibit self-written simulations. Simulation tools supplies execution support but does not independently organize or enforce coverage of every required function. Full harness and binary feedback share the requirement-level workflow, differing in the diagnostic simulation feedback returned to the constructor.

\begin{table}[!htbp]
\centering
\small
\caption{Model-visible support in the four harness settings.}
\label{tab:comparison-condition-materials}
\begin{tabularx}{\linewidth}{@{}lY@{}}
\toprule
Setting & Model-visible materials and feedback \\
\midrule
Requirements only & Functional requirements and ordinary construction tools; agents may implement their own verification. \\
Simulation tools & The same requirements, with the simulation environment, starter code, and API guidance; agents select and interpret their own tests. \\
Full harness & Simulation support, independent requirement-level test binding, and diagnostic execution feedback throughout revision. \\
Binary feedback & The same harness workflow, with pass/fail simulation feedback in place of diagnostic details. \\
\bottomrule
\end{tabularx}
\end{table}

\paragraph{Requirements-only setting.}
This setting follows ordinary user interaction with an agent: the user describes the desired mechanism and its functions in a natural-language request, and the agent generates a corresponding CAD assembly. The agent chooses how to construct the mechanism and verify its functionality, without requirement-level testbeds or diagnostic feedback from \harnessname.

\section{Mechanism Families and Task Taxonomy}
\label{app:mechanism-catalog}

\subsection{Mechanism Catalog}

\benchname{} comprises the 46 mechanism families listed in
Table~\ref{tab:mechanism-families}, numbered consecutively from 1 to 46.
Each family is defined by its required functions and fixed constraints,
while allowing different geometries, transmission arrangements, and
permitted operating choices.
Different designs that satisfy the requirements of a family are treated
as alternative realizations rather than additional families.
The table summarizes each family's functional requirements and the
operations and observations used to evaluate them.
The specific components, inputs, measurements, and operating parameters
are established through independent binding to each candidate design.

\begingroup
\small
\setlength{\tabcolsep}{5pt}
\setlength{\arrayrulewidth}{0.3pt}
\renewcommand{\arraystretch}{1.12}
\setlength{\LTleft}{0pt}
\setlength{\LTright}{0pt}
\setlength{\LTcapwidth}{\linewidth}
\arrayrulecolor{black!22}

\begin{longtable}{@{}
    >{\raggedright\arraybackslash}
        p{\dimexpr0.24\linewidth-\tabcolsep\relax}
    >{\raggedright\arraybackslash}
        p{\dimexpr0.44\linewidth-2\tabcolsep\relax}
    >{\raggedright\arraybackslash}
        p{\dimexpr0.32\linewidth-\tabcolsep\relax}
@{}}

\caption{The 46 mechanism families in \benchname{}, their functional
requirements, and the corresponding operations and observations.}
\label{tab:mechanism-families}\\
\hline
\rowcolor{black!6}
\textbf{Mechanism} &
\textbf{Functional requirements} &
\textbf{Operations and observations} \\
\hline
\endfirsthead

\multicolumn{3}{@{}l}{\textit{\tablename~\thetable{} continued}}\\[5pt]
\hline
\rowcolor{black!6}
\textbf{Mechanism} &
\textbf{Functional requirements} &
\textbf{Operations and observations} \\
\hline
\endhead

\multicolumn{3}{r@{}}{\textit{Continued on next page}}\\
\endfoot

\endlastfoot

\textbf{1}\newline
Hinged toolbox &
Open the lid to the target angle against a physical stop, then close it;
keep the lid attached and the hinge pin axially retained throughout. &
Actuate the handle or lid edge; record lid angle, stop contact, and
axial hinge-pin displacement. \\[5pt]
\hline

\textbf{2}\newline
Rack-and-pinion gate &
Raise and lower the gate, including under load, with the required travel
per handle revolution while maintaining guidance and respecting travel
limits. &
Turn the handle; record input rotation, gate travel, load, guide
retention, and travel limits. \\[5pt]
\hline

\textbf{3}\newline
Four-bar wiper &
Convert continuous input rotation into repeated sweeping with the
required angular range, consistent return, and intact linkage
connections. &
Rotate the drive shaft; record sweep extrema, return position, cycle
count, and accumulated drift. \\[5pt]
\hline

\textbf{4}\newline
Hand-cranked egg beater &
Drive both beaters in opposite directions at the required speed ratio,
including under the specified resistance. &
Turn the handle; record handle and beater angles and speeds, resisting
loads, and shaft retention. \\[5pt]
\hline

\textbf{5}\newline
Self-locking utility knife &
Extend the blade to two positions, retain each position after release,
and permit unlocking and full retraction. &
Operate the slider; record extension at each position, displacement
after release, and the final tip position. \\[5pt]
\hline

\textbf{6}\newline
Folding drying rack &
Unfold both frames and raise the drying surface, stand stably, and fold
into the required storage configuration without losing support. &
Operate the folding controls; record frame angles, drying-surface
height, ground support, and folded dimensions. \\[5pt]
\hline

\textbf{7}\newline
Bicycle &
Provide pedal-driven travel, steering, and bell operation; cargo
variants additionally require loaded travel and steering. &
Operate the pedals, handlebar, and bell; record drive ratio, wheel
contact, stability, and cargo loading where applicable, and assess bell
sound separately. \\[5pt]
\hline

\textbf{8}\newline
Pedal-operated bin &
Open and hold the lid while the pedal is pressed, then close the lid
and return the pedal after release over repeated cycles. &
Press and release the pedal; record lid angle, pedal return, and closure
after release. \\[5pt]
\hline

\textbf{9}\newline
Mechanical alarm clock &
Run after winding, display and adjust time, retain an independently set
alarm, and trigger and stop the alarm without interrupting timekeeping. &
Operate winding, time-setting, alarm-setting, and stop controls; record
hand motion, settings, and striking, with sound assessed through a
separate physical test. \\[5pt]
\hline

\textbf{10}\newline
Kitchen tongs &
Grasp an object, retain it while the tongs are lifted, and reopen on
release to return the object to the table. &
Squeeze and lift the tongs; record jaw gap, object height and slip, and
table contact. \\[5pt]
\hline

\textbf{11}\newline
Retractable ballpoint pen &
Complete two extension--retraction cycles, with successive button
presses and releases alternating and retaining the tip state. &
Press and release the button; record press count, released button
position, and tip state. \\[5pt]
\hline

\textbf{12}\newline
Nail clipper &
Repeatedly close the cutting edges to the required gap and reopen them
when the lever is released. &
Operate the lever; record edge separation, reopening, and pressing
force; material cutting requires a separate physical test. \\[5pt]
\hline

\textbf{13}\newline
One-way ratchet screwdriver &
Accumulate forward output rotation under reciprocating handle motion,
both unloaded and under the specified resisting torque. &
Turn the handle back and forth; record net output rotation, reverse
motion during return, and resisting torque. \\[5pt]
\hline

\textbf{14}\newline
Linearly actuated scissor lift &
Raise and lower the platform within the input-force limit while
maintaining its orientation, guidance, and support, and hold its height
under load. &
Operate the lift input; record platform height and tilt, applied load,
driving force, and holding drift. \\[5pt]
\hline

\textbf{15}\newline
Equal-arm balance &
Balance equal loads, tilt toward the heavier side, recover after
removing excess load, retain freely suspended pans, and indicate mass
difference. &
Load and unload the pans; record mass difference, beam tilt, pan pose,
pointer reading, and settling behavior. \\[5pt]
\hline

\textbf{16}\newline
Toggle clamp / press &
Close and reopen using the selected variant's controls; self-locking
variants must resist an opening load, while pressing variants must
repeat loaded strokes and return. &
Operate the handle or release control; record closing travel, clamping
or pressing force, loaded drift, and return. \\[5pt]
\hline

\textbf{17}\newline
Dry piston-pump mechanism &
Repeat piston strokes with the required travel and return position,
both unloaded and under resistance, while maintaining guidance and
connections. &
Rotate the crank; record piston position, stroke, period, and
resistance; the task evaluates dry motion rather than fluid pumping.
\\[5pt]
\hline

\textbf{18}\newline
Three-digit combination lock &
Adjust each wheel independently, unlock with the correct code, re-lock
after reinsertion and scrambling, and reject a single wrong digit at
each position. &
Operate the wheels and bolt; record code settings, locking-bar or bolt
travel, and blocking under incorrect codes. \\[5pt]
\hline

\textbf{19}\newline
Two-digit mechanical counter &
Count correctly through \(09\rightarrow10\) and \(19\rightarrow20\),
increment once per forward stroke, prevent reverse counting on return,
and retain the reading at rest. &
Operate the push rod; record the units and tens readings after each
forward stroke, return stroke, and release. \\[5pt]
\hline

\textbf{20}\newline
Three-ball dispenser &
Dispense exactly three balls in each of three cycles, reaching
cumulative totals of 3, 6, and 9; refill between cycles while retaining
unused balls and conserving the total count. &
Operate the dispensing slider; track individual balls, counts, and
cycle timing across the hopper, metering chamber, and receiving area.
\\[5pt]
\hline

\textbf{21}\newline
Gravity timer &
Run autonomously after release, trigger within the specified time
window without triggering early, retain a visible indicator, and repeat
after reset. &
Operate reset and release; record weight position, trigger time, and
indicator retention. \\[5pt]
\hline

\textbf{22}\newline
Slider-crank push rod &
Complete three unloaded reciprocating cycles with the required forward
and return travel, then repeat under the specified load. &
Rotate the crank; record slider position, cycle count, stroke length,
and applied load. \\[5pt]
\hline

\textbf{23}\newline
Gravity-pawl turntable &
Allow forward rotation with gravity-driven pawl re-engagement, block
reverse rotation under the specified torque, and resume forward
rotation. &
Rotate the turntable; record forward and reverse angles, pawl seating
position and time, and reverse torque. \\[5pt]
\hline

\textbf{24}\newline
Sliding-bolt gate &
Remain locked under pushing with the bolt inserted, then open to the
required angle after withdrawal while retaining hinge support and bolt
guidance. &
Operate the bolt and gate handle; record bolt travel, opening angle,
and gate motion while locked. \\[5pt]
\hline

\textbf{25}\newline
Three-stage gear reducer &
Transmit rotation at \(3{:}1\) per stage and \(27{:}1\) overall, operate
under three braking loads, and decelerate naturally after input removal. &
Drive only the input shaft; record each shaft's angle and speed, input
energy, and braking dissipation. \\[5pt]
\hline

\textbf{26}\newline
Desktop display turntable &
Rotate to \(90^\circ\pm3^\circ\) and return within \(\pm3^\circ\) of the
initial position without eccentric motion or axial separation. &
Apply bounded force to the manual tab; record rotation angle, return
error, eccentricity, and axial displacement. \\[5pt]
\hline

\textbf{27}\newline
Small swing gate &
Open to \(60^\circ\pm3^\circ\) and return within \(\pm3^\circ\) of the
initial position while maintaining a stable frame and preventing
hinge-axis slip or disengagement. &
Push and pull the gate; record opening angle, return error, and downward
displacement along the hinge axis. \\[5pt]
\hline

\textbf{28}\newline
Tilting desktop mirror &
Tilt \(30^\circ\pm3^\circ\) from upright and return within
\(\pm3^\circ\) of the initial position while maintaining base stability
and pivot support. &
Rotate the mirror frame; record tilt and axial displacement; angle
locking is not required. \\[5pt]
\hline

\textbf{29}\newline
Folding wall hook &
Deploy the hook through \(90^\circ\pm3^\circ\) from its stowed position
and return within \(\pm3^\circ\) while retaining the fixed base and
hinge connection. &
Operate the hook arm; record rotation angle, return error, and axial
hinge displacement. \\[5pt]
\hline

\textbf{30}\newline
Spring clothespin &
Open when squeezed, retain the specified test piece after release,
release it when squeezed again, and repeatedly return to the closed
position. &
Squeeze and release the tails; record jaw gap, test-piece slip, release,
and the specified load. \\[5pt]
\hline

\textbf{31}\newline
Scissors &
Repeatedly open and close through the required angular range while
retaining the pivot and avoiding blade interference. &
Operate the handles; record blade angle, opening, and pivot retention;
opening and closing do not establish material-cutting capability.
\\[5pt]
\hline

\textbf{32}\newline
Ratchet hand winch &
Wind under load, hold during handle return or reverse drum torque,
permit controlled payout, and re-engage to hold and wind again. &
Operate the handle and release control; record drum rotation, load
displacement, and drift after stopping. \\[5pt]
\hline

\textbf{33}\newline
Spring-latch door lock &
Retract the latch when the handle turns, extend it on release, and
automatically retract and re-engage against the strike during door
closure. &
Operate the handle and close the door; record latch travel, door
position, and contact with or entry into the strike. \\[5pt]
\hline

\textbf{34}\newline
Reversible dual-pawl ratchet wrench &
Drive the head in the selected direction without reversing it on the
return stroke; switch direction and repeat, including under resistance. &
Operate the handle and selector; record output advance, reverse motion
on return, and resisting torque in both modes. \\[5pt]
\hline

\textbf{35}\newline
Full-cycle ratchet crimper &
Reach the required loaded stroke and force, prevent opening after a
partial stroke, and release for full reopening at the endpoint over
repeated cycles. &
Operate the handles; record head travel, force, and pawl state; metal
forming and electrical-joint quality are outside this check. \\[5pt]
\hline

\textbf{36}\newline
Geneva indexing table with overload clutch &
Index by \(90^\circ\) with stable dwell under load, resist dwell
disturbances, slip under obstruction, and resume correct indexing after
the obstruction is removed. &
Drive the input and apply disturbances or obstruction; record table
angle, dwell drift, rotations on both sides of the clutch, and load.
\\[5pt]
\hline

\textbf{37}\newline
Folding shopping cart &
Carry the specified load while rolling and permit handle folding and
unfolding without losing assembly integrity. &
Operate the pulling and handle-folding controls; record travel, load,
frame support, wheel motion, and handle angle. \\[5pt]
\hline

\textbf{38}\newline
Mechanical \(4\times4\) puzzle cube &
Rotate and align outer and inner layers separately without unintended
motion of other layers or piece loss, and execute successive turns
about different axes. &
Turn the selected layers; record inner and outer layer angles,
unintended layer motion, and piece retention. \\[5pt]
\hline

\textbf{39}\newline
Twin-cylinder engine linkage &
Drive both pistons through repeated strokes with the required phase
relationship and travel while maintaining guidance and avoiding
interference. &
Rotate the crankshaft; record piston positions, phase, and stroke;
the task evaluates transmission rather than combustion. \\[5pt]
\hline

\textbf{40}\newline
Four-station indexing table &
Cycle through the designated stations under continuous input,
alternating motion and dwell without disengagement or jamming. &
Rotate the input shaft; record table angle, stopping positions, and
dwell duration. \\[5pt]
\hline

\textbf{41}\newline
Synchronized parallel gripper &
Open and close symmetrically from a single input while maintaining
parallel jaw faces and guidance, and grasp and retain a test piece
under load; release and repeat the grasp--hold--release sequence. &
Operate the gripper; record jaw travel, face angle, gripping force,
test-piece slip, and the state left for the next grasp. \\[5pt]
\hline

\textbf{42}\newline
Releasable reduction winch &
Lift under load through a reduction transmission, hold against reverse
drum torque after input stops, permit controlled payout, and re-engage
for holding and winding. &
Operate the handle and release control; record input and drum rotation,
load displacement, holding drift, and release and re-locking behavior.
\\[5pt]
\hline

\textbf{43}\newline
Baby stroller &
Roll and steer under load, engage and release the brake, and, for
folding variants, fold, unfold, and re-lock the frame. &
Operate the push handle, steering, brake, and folding controls; record
body pose, wheel motion, load support, and residual displacement after
braking. \\[5pt]
\hline

\textbf{44}\newline
Height- and tilt-adjustable worktable &
Reach the required height and tilt ranges, hold both settings under
load, and permit unlocking, readjustment, and re-locking. &
Operate the lift, tilt, and locking controls; record height, angle,
load, and drift after locking. \\[5pt]
\hline

\textbf{45}\newline
Coin sorter and counter &
Sort a continuous mixed coin stream to the correct outlets, count each
coin exactly once, and resume after pauses or collection-bin emptying
without count errors. &
Operate the feeder; record coin type and size, outlet assignment,
outlet counts, counter readings, and jams. \\[5pt]
\hline

\textbf{46}\newline
Interchangeable-tune music box &
Play and stop in the required direction and rhythm, actuate sound
elements in sequence with recovery, and accept a replacement tune
component. &
Operate power or winding, stop, and tune replacement; record score
position and actuation sequence, with melody recognition assessed
through a separate physical test. \\[5pt]
\hline

\end{longtable}
\endgroup

\subsection{Taxonomy Dimensions and Representative Mechanisms}

We characterize the mechanical demands of the tasks along five
complementary dimensions: motion primitives, topology, complexity,
contact, and geometric difficulty.
These dimensions are not mutually exclusive categories: a mechanism
family typically involves several dimensions, and task variants may
place different demands on each.

\paragraph{Motion primitives.}
This dimension describes how a mechanism converts and transmits motion,
including input--output direction, phase relationships, and repeated or
intermittent operation.
In the four-bar wiper (3), continuous crank rotation must produce
repeated wiper oscillation, so evaluation must capture the input--output
relationship over complete motion cycles.

\paragraph{Topology.}
This dimension describes connections, branches, and closed kinematic
chains, together with the compatibility of the constraints they impose.
In the scissor lift (14), revolute and prismatic connections form closed
loops that constrain platform motion; the connections must therefore be
individually valid and collectively kinematically compatible.

\paragraph{Complexity.}
This dimension describes interactions among components and subsystems,
state transitions, required operating sequences, and part count.
The mechanical alarm clock (9) couples energy supply, timekeeping, and
alarm operation through winding, running, and triggering, requiring
coordination across subsystems and states.

\paragraph{Contact.}
This dimension describes how engagement, sliding, locking, friction,
and changes in contact state enable or prevent a behavior.
In the self-locking utility knife (5), pressing the release changes
the contact state to permit sliding, after which engagement at another
detent restores locking.

\paragraph{Geometric difficulty.}
This dimension describes the feature construction, alignment, clearance
control, and assembly access needed to realize the intended mechanical
relationships.
The reversible dual-pawl ratchet wrench (34) illustrates these demands
through its pawl--tooth geometry, pivot fit, and local clearances, whose
shapes and relative placement must support the intended engagement.

The five dimensions interact: adding components may introduce new
contacts, while changing the topology may simplify geometry but
complicate operation.
The taxonomy therefore describes the mechanical diversity of the
benchmark, while performance on an individual family generally reflects
the combined effects of several capabilities.

\clearpage
\section{Functional Specifications and Diagnostic Cases}
\label{app:functional-specifications}

We use Fable's design trajectory for the synchronized parallel gripper
(Task~41) to illustrate how functional requirements guide construction,
execution, and revision, and how the resulting evidence establishes
completion.
Appendix~\ref{app:repair-trajectories} adds two repair cases that connect
functional feedback to local geometric changes and renewed verification.

\subsection{From Input Requirements to Executable Tests}
\label{app:requirement-test-correspondence}

Table~\ref{tab:requirement-test-correspondence} connects three levels of
specification for the cases below. The \textit{input requirement} states
the intended behavior. The \textit{functional test template} decomposes
that behavior into operations, observations, and acceptance checks,
with interfaces and parameters to be bound to a candidate. The
\textit{candidate-specific simulation test} supplies the actual controls,
environment, measurements, and declared parameter values.
Registered tasks reuse the template across candidates; open-ended
requests require a new template and independent review, as described in
Appendix~\ref{app:prompts}.

\begingroup
\small
\setlength{\tabcolsep}{5pt}
\renewcommand{\arraystretch}{1.10}
\setlength{\LTleft}{0pt}
\setlength{\LTright}{0pt}
\setlength{\LTcapwidth}{\linewidth}
\begin{longtable}{@{}
>{\raggedright\arraybackslash}p{\dimexpr0.24\linewidth-\tabcolsep\relax}
>{\raggedright\arraybackslash}p{\dimexpr0.33\linewidth-2\tabcolsep\relax}
>{\raggedright\arraybackslash}p{\dimexpr0.43\linewidth-\tabcolsep\relax}@{}}
\caption{From input requirements to functional test templates and recorded
candidate-specific simulation tests. Prompt clauses are condensed;
numerical settings refer to the named candidate and revision.}
\label{tab:requirement-test-correspondence}\\
\toprule
\textbf{Input requirement} & \textbf{Functional test template} &
\textbf{Candidate-specific simulation test} \\
\midrule
\endfirsthead
\multicolumn{3}{@{}l}{\textit{\tablename~\thetable{} continued}}\\[4pt]
\toprule
\textbf{Input requirement} & \textbf{Functional test template} &
\textbf{Candidate-specific simulation test} \\
\midrule
\endhead
\midrule
\multicolumn{3}{r@{}}{\textit{Continued on next page}}\\
\endfoot
\bottomrule
\endlastfoot

\multicolumn{3}{@{}l}{\textbf{Task 41: synchronized parallel gripper --- Fable, r02}}\\*[4pt]
A single input opens and closes both jaws with symmetric motion and
parallel gripping faces. &
\textit{Motion and guidance.} Operate the normal input; check each jaw's
travel, synchronization, face parallelism, guide deviation, and available
opening. &
Apply force to the central plunger along its sliding axis, capped at
35\,N. Measure the actual jaws: each travels at least 6\,mm;
synchronization and guide errors stay within 0.5\,mm, parallelism within
0.01\,rad, and the released opening reaches at least 50\,mm. \\[5pt]
Hold the test piece without slipping or falling under the specified load. &
\textit{Loaded holding.} Apply a load to the grasped piece; verify actual
load, minimum jaw--piece contact force, and three-dimensional slip
relative to the jaws. &
Apply 5\,N tangentially to the free test block during 1--3 and 5--7\,s.
Require at least 10\,N grip force, measured mean load of at least 4.5\,N,
and slip of at most 1\,mm in each loaded phase. \\[5pt]
Grasp, release, and repeat using the same operating input. &
\textit{Repeated use.} Execute two close--hold--release cycles in one
trial, retaining the workpiece state between cycles. Apply the same
checks to both cycles. &
The \texttt{repeated\_loaded\_grip} trial lasts 8\,s. Plunger targets are
+40\,mm for closing and 0\,mm for opening; physical contact limits actual
travel. The second cycle begins at 4\,s without resetting the block.
All 16 checks are evaluated on this continuous trial. \\[5pt]
\midrule

\multicolumn{3}{@{}l}{\textbf{Task 33: spring-return door latch --- Grok, r06}}\\*[4pt]
Turning the handle retracts the latch so the door can open. &
\textit{Unlocking and opening.} Retract the latch through handle
operation before opening the door; observe latch protrusion and door
position. &
Bind inputs to the actual handle and door hinges, with torque caps of
3 and 12\,N\,m. Target a handle angle of $-70^\circ$ and a door opening of
0.4\,rad. Require latch protrusion of at most 8\,mm and door-position
error of at most 0.12\,rad. \\[5pt]
Releasing the handle lets the latch extend automatically. &
\textit{Passive return.} Remove handle actuation while the door is open;
check that the spring restores latch extension. &
After opening, stop driving the handle. Read the actual nose position
relative to the door and require at least 10\,mm protrusion at the end
of the return phase. \\[5pt]
Closing compresses the beveled latch, which springs into the strike
plate and locks again; repeat the sequence. &
\textit{Contact-mediated re-locking.} Close the door; check strike
contact, temporary latch retraction, closed position, and engagement
after door actuation ends. Repeat the full sequence. &
The \texttt{repeated\_open\_close} trial uses 0.8\,s phases and two cycles
over 8.8\,s. Measure strike-contact force and exported latch-nose
penetration into the actual strike opening. Require contact, protrusion
of at most 8\,mm during closure, and positive final engagement in both
cycles. \\[5pt]
\midrule

\multicolumn{3}{@{}l}{\textbf{Task 39: twin-cylinder linkage --- Grok, r02}}\\*[4pt]
Continuous crankshaft rotation drives both pistons repeatedly with the
design's strokes and phase relationship. &
\textit{Transmission and phase.} Drive the crank; measure both piston
trajectories, complete reciprocation counts, and crank angles at matched
piston extrema. &
The \texttt{continuous\_crank\_drive} trial drives the shaft for 3.5\,s
at a target of $2\pi$\,rad/s with a 1\,N\,m torque cap. Require at least
two cycles per piston, 24\,mm stroke within 2\,mm, and phase separation
of $\pi$ within 0.1\,rad over at least two matched extrema pairs. \\[5pt]
Both pistons stay within their guides without unintended interference
or jamming. &
\textit{Guidance and interference.} Monitor guide deviation and
unintended contact throughout the same crank-driven trial; distinguish
normal bearing contacts from prohibited collisions. &
Track both ends of each piston skirt in the frame coordinates; require
guide error of at most 1\,mm. Sum contact forces outside the identified
bearing regions and require at most 1\,N. Together with the motion and
phase checks, the trial evaluates eight functional checks. \\
\end{longtable}
\endgroup

The mapping is not one requirement per simulation run: several checks
can share a continuous trial, preserving the states needed for later
operations. For example, the second grasp uses the block left by the
first release, and re-locking is checked after closing inputs have been
removed. The trials also undergo source-geometry auditing at sampled
poses, which is separate from the functional checks summarized above
(Appendix~\ref{app:physics}). These protocols test the required
behaviors for particular designs; a different mechanism may need a
different input location, operating policy, or observation mapping.

\clearpage

\subsection{Worked Example: Designing and Refining a Parallel Gripper}
\label{app:gripper-trajectory}

\begin{figure}[!htbp]
\centering
\includegraphics[width=\linewidth]{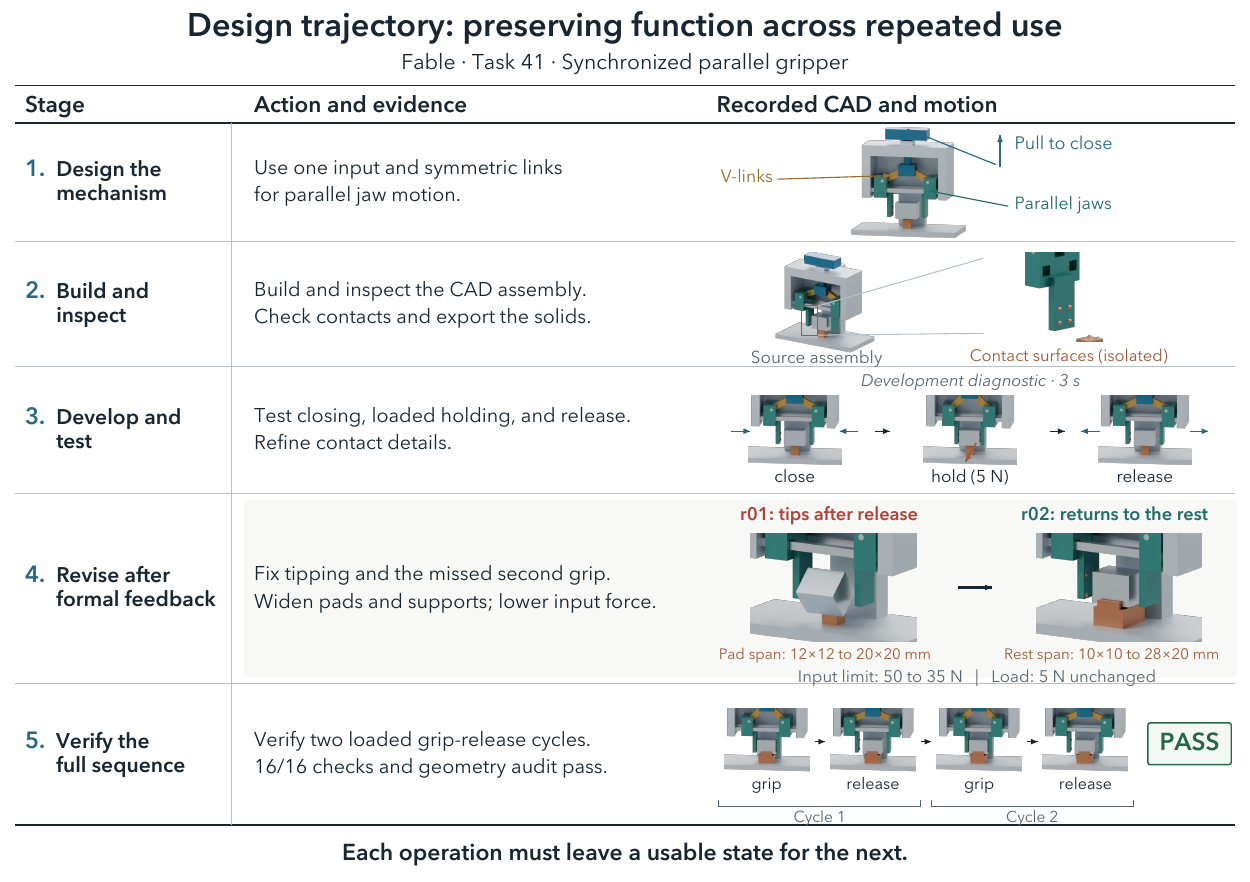}
\caption{Fable's design trajectory for the synchronized parallel gripper
(Task~41). Construction and development tests establish the initial
grasping behavior; formal feedback exposes a failure during repeated
use. The agent revises the contact geometry, workpiece support, and
permitted input effort before the final candidate passes both loaded
grasp--release cycles and the source-geometry audit.}
\label{fig:task41-design-trajectory}
\end{figure}

The task requires a single input to drive two synchronized, parallel
jaws that grasp a workpiece, hold it under load, release it, and repeat
the operation.
Fig.~\ref{fig:task41-design-trajectory} follows the agent from choosing
a linkage and inspecting its CAD assembly to developing tests,
interpreting formal feedback, and verifying the revised design.

The first submitted design completes the initial loaded grasp, but its
revision report identifies a subsequent failure: after release, the
workpiece tips off its rest, leaving the jaws to close on empty space
in the second cycle.
The agent widens the contact pads and support footprint and lowers the
permitted input effort, while retaining the external test load.
The revised candidate passes the complete sequence.
This trajectory illustrates why a successful operation must leave the
mechanism and its environment in a usable state for the next operation.

\clearpage
\subsection{Fixed Requirements, Adjustable Parameters, and Binding}
\label{app:requirements-binding}

The specification distinguishes \textit{functional requirements},
\textit{fixed constraints}, and \textit{adjustable operating parameters}.
For Task~41, synchronized parallel motion, loaded holding, release,
and repeated use remain required throughout revision.
The single-input constraint also remains fixed.
As a registered benchmark task, it uses a frozen test template;
\textit{candidate-specific binding} makes that template executable for
the submitted mechanism.

\paragraph{Candidate-specific binding.}
The final evaluation agent connects the template to the candidate's
\textit{mechanism}, \textit{environment}, \textit{control policy},
\textit{observations}, and \textit{pass criteria}.
The mechanism uses a central plunger, symmetric links, and guided jaws
to grasp a free workpiece resting on a support; an external load is
applied during holding.
The policy pulls the plunger to close the jaws, holds under load,
releases, and repeats.
Observations include jaw synchronization and parallelism, grip force,
workpiece slip, and its state after release.
The pass criteria require these measurements to satisfy the task's
motion and holding requirements throughout the sequence.

\paragraph{Operating protocol.}
Different gripper designs may place the input, contact surfaces, and
supports differently, so the locations and directions of actuation,
load application, and measurement must follow the candidate.
The constructor's operating instructions inform these choices, while
the test-binding agent checks them against the delivered geometry and mechanics.
Binding changes how the required behavior is exercised and observed;
it preserves the behaviors that must be established.

\paragraph{Parameter provenance and permitted adjustments.}
Each parameter retains its value and supporting basis, whether derived
from the task, source documentation, geometric measurements, or an
explicit constructor decision.
A permitted adjustment records the original target, revised target,
and reason for the change, following Appendix~\ref{app:permitted-adjustments}.
In this trajectory, the input-effort limit changes from 50 to 35\,N,
while the external 5\,N load and the requirements for loaded holding,
release, and repeated use remain unchanged.
The final candidate is evaluated with the 35 N input-effort limit, while the external 5 N load and the requirements for loaded holding, release, and repeated use remain unchanged.

\subsection{State Dependencies and Completion}
\label{app:state-completion}

Each trial starts from a fresh instance of the delivered configuration,
while successive operating phases preserve the evolving physical state.
In the gripper test, opening the jaws releases the workpiece without
resetting its position before the next grasp.
An isolated successful grasp therefore does not establish repeatability:
the released workpiece must remain accessible to the mechanism.

\textit{Whole-task completion} requires every required function and the
applicable physical-validity checks to pass under the declared operating
conditions and modeling assumptions.
For the final gripper, this includes both loaded grasp--release cycles
and the source-geometry audit.
All evidence must refer to the same delivered configuration, which
remains unchanged during evaluation.
After a revision, the binding and evaluation must be checked against the
updated candidate; successful checks from different designs cannot be
combined to establish completion of one design.

\subsection{Diagnostic Interpretation}

The gripper trajectory illustrates how feedback separates verified
behavior, observed failures, and missing evidence.

\paragraph{Partial functional evidence.}
Development tests of closing, loaded holding, and release provide
useful evidence for refining contacts, but do not by themselves establish
that the complete operating sequence succeeds.
Likewise, a passed first grasp leaves the second grasp unverified.

\paragraph{Observed functional failure.}
In the first submission, the second closure fails to grip the workpiece.
The failed grip and holding checks identify the unmet behavior;
examining the preceding release explains why that behavior is no longer
possible from the resulting state.
Repair therefore involves the workpiece support and release behavior,
in addition to the geometry that produces the grasp.

\Needspace{5\baselineskip}
\paragraph{Physical-validity failure.}
During development of the revision, a flat open-stop contact triggers
a source-geometry overlap check and is replaced with three small
support buttons.
A desired motion can therefore coexist with a geometric violation.
Functional measurements and source-geometry findings provide distinct
feedback and both contribute to the completion decision.

\paragraph{Insufficient evidence.}
If a required operation is not executed or its necessary observation
is unavailable, the corresponding function remains inconclusive.
For example, omitting the second loaded grasp would leave repeatability
unestablished even if the first cycle passed.
Missing evidence establishes neither mechanical failure nor completion.

\paragraph{Test outcomes and defect attribution.}
A failed grip check identifies an unmet criterion, but does not by
itself distinguish inadequate contact force from a misplaced workpiece.
That distinction requires evidence about contacts and the preceding
motion; here, the revision report links the missed second grasp to
post-release tipping.
When assessing diagnostic correctness, reference defect labels must
be established independently of the test outcome being evaluated.

\Needspace{32\baselineskip}
\subsection{Repair Trajectories}
\label{app:repair-trajectories}

The following cases show how functional feedback and geometry checks
guide successive revisions of an existing design.

\paragraph{Restoring re-locking after closure (Task~33).}
Fig.~\ref{fig:task33-relocking-repair} follows Grok's repair of a
spring-return door latch.
In revision r04, the door opens successfully, but the latch rests on the
strike-plate edge after closing instead of entering its opening.
Both re-locking checks fail, leaving 13 of 15 functional checks passed.
Grok reshapes the entry bevel and adds a catching shoulder; revision r05
restores engagement after both closures and passes all 15 functional
checks.
A remaining latch--faceplate interference is then removed through a
local faceplate relief in r06, which also passes the source-geometry
audit.
Test parameters remain unchanged throughout these revisions.
The case distinguishes reaching a closed door position from restoring
the latch engagement required for re-locking, and separates functional
recovery from subsequent geometric validation.

\begin{figure}[!htbp]
\centering
\includegraphics[width=\linewidth]{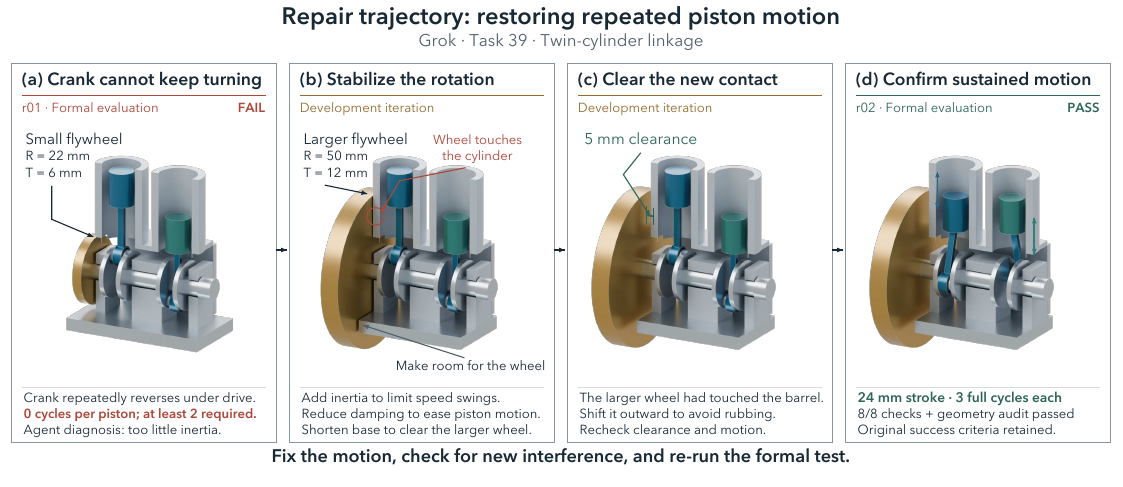}
\caption{Grok repairs the twin-cylinder linkage (Task~39): enlarge the flywheel, clear the resulting cylinder contact, and verify repeated piston motion.}
\label{fig:task39-piston-repair}
\vspace{-2mm}
\end{figure}
\FloatBarrier

\Needspace{29\baselineskip}
\paragraph{Restoring sustained piston motion (Task~39).}
Fig.~\ref{fig:task39-piston-repair} follows Grok's repair of a
twin-cylinder linkage.
In revision r01, the crank repeatedly reverses under drive and neither
piston completes a full cycle.
The agent diagnoses insufficient rotational inertia and enlarges the
flywheel, while reducing guide damping and shortening the bedplate.
Intermediate development checks then reveal interference between the
larger flywheel and a cylinder barrel.
Moving the wheel outward provides approximately 5\,mm of clearance.
In the formal evaluation of r02, both pistons complete three cycles
with approximately 24\,mm stroke; all eight functional checks and the
source-geometry audit pass under the original success criteria.
This case shows how a repair aimed at sustained motion can introduce
a new geometric conflict, requiring clearance checks followed by
verification of the complete behavior.

\FloatBarrier

\section{Scene Interface and Physical Execution Details}
\label{app:physics}

\subsection{Scene Interface and Initialization}
\label{app:source-scene}

The physical scene associates the submitted geometry with rigid bodies,
connections, material properties, external supports, and gravity.
Source preparation constructs the simulation model from these geometric
and physical declarations before test binding.
The binding connects the template's operating inputs and observation
signals to the prepared model.
Prechecks verify that these interfaces are available and that the source
bodies and their physical-update routines are retained.

During execution, prescribed inputs are applied at the bound operating
interfaces, subject to the force or torque limits specified by the test
protocol.
Trial initialization and state continuity follow the rules in
Appendix~\ref{app:state-completion}.
Parameter provenance and permitted adjustments are recorded as described
in Appendix~\ref{app:requirements-binding}.

\subsection{Contact, Connections, and Equivalent Models}
\label{app:contact-connections}
\label{app:equivalent-models}

The registered-template execution path uses the NSC system in
Project Chrono \citep{Chrono2016} for rigid-body dynamics
and Bullet for collision detection.
Collision representations are constructed from the retained source
geometry, and the material and contact parameters used during execution
are recorded with the run.

The supported connection types are revolute, prismatic, spherical,
and fixed.
Each connection is associated with finite regions on the source geometry
and remains active only while those regions overlap.
If the regions separate, the connection is disabled before the next
integration step and remains released for the rest of the trial.

Selected mechanical effects, including spring storage, elastic response,
friction-clutch behavior, detents, and gear engagement, can be represented
by explicitly declared equivalent models.
These models apply force or torque laws to the corresponding source
components while the native solver computes the resulting rigid-body
motion.
Their parameters, source correspondence, and operating ranges are
specified before execution.

The original CAD geometry is retained, and any exclusions from
motion-geometry auditing are restricted to declared elastic reference
bodies or explicitly bounded local regions.
Other geometry and contacts remain subject to the applicable checks.

\begin{table}[t]
\caption{Task~25 reducer timings (1\,ms step; standard test budget: 300\,s).
Different designs and simulated durations preclude a controlled speedup comparison.}
\label{tab:simcost}
\centering
\small
\setlength{\tabcolsep}{5pt}
\renewcommand{\arraystretch}{1.0}
\begin{tabularx}{\linewidth}{@{}>{\raggedright\arraybackslash}X rrr@{}}
\toprule
Interaction model & Wall/step (s) & Sim. time (s) & Total wall (s) \\
\midrule
Native tooth contact (light)
  & 0.894 & 6.0 & \textbf{5,364} \\
Equivalent spur gears (light/medium/heavy)
  & 0.113--0.117 & 1.2 & \textbf{136--141} \\
\bottomrule
\end{tabularx}
\end{table}

\subsection{Trajectory Recording and Geometry Auditing}
\label{app:trajectory-auditing}

The primary registered-template path records measurement signals at
every native integration step.
These signals include the motion, connection states, and contact
information required by the bound tests.
Body poses are saved at nominal 20\,ms intervals, with additional poses
retained at operating-phase boundaries.

Functional assertions are evaluated from the recorded measurement
signals, while source-geometry auditing uses the saved body poses.
The audit selects poses at 10\,Hz, includes the initial and final poses,
and checks the original source geometry for violations of the applicable
geometric criteria.
The 20\,ms recording interval determines the retained motion detail,
whereas the 10\,Hz audit frequency determines when the geometry is
inspected.
Geometric conclusions apply to the inspected poses within the declared
modeling scope.

If a trial fails or is interrupted, the measurements and poses recorded
up to that point are retained together with its execution status.

\section{Agent Configurations, Budgets, and Evaluation Protocol}
\label{app:evaluation-protocol}

\subsection{Model and Agent Configurations}
\label{app:agent-configurations}

Table~\ref{tab:agent-configurations} lists the model identifiers, agent wrappers, and requested reasoning settings used during generation and final evaluation. During generation, each agent uses its own model configuration. Final batch evaluation uses GPT-6 Astra with a one-million-token context allowance and medium reasoning across all experimental settings.

\begin{table}[t]
\centering
\small
\setlength{\tabcolsep}{4pt}
\renewcommand{\arraystretch}{1.15}
\caption{
Generation-agent and final-evaluation configurations
}
\label{tab:agent-configurations}
\begin{tabular}{@{}llll@{}}
\hline
\textbf{Role} &
\textbf{Model identifier} &
\textbf{Wrapper} &
\shortstack[l]{\textbf{Requested}\\\textbf{reasoning setting}} \\
\hline
Constructor &
\texttt{claude-fable-5-1-thinking-high} &
Cursor &
High \\

Constructor &
\texttt{cursor-grok-4.6-xhigh} &
Cursor &
Extra high \\

Constructor &
\texttt{kimi-code/k3} &
Kimi CLI &
High\textsuperscript{*} \\

Constructor &
\texttt{gpt-6-astra} &
Codex &
High \\

Final evaluation agent &
\texttt{gpt-6-astra} &
Codex &
Medium \\
\hline
\end{tabular}

\smallskip
\begin{minipage}{\linewidth}
\footnotesize
\textsuperscript{*}For Kimi CLI, High records the requested setting;
the integration does not expose a separate reasoning-effort control.
\end{minipage}
\end{table}

Both Fable 5.1 and GPT-6 Astra are configured with a context allowance
of one million tokens.
Each constructor receives the registered English task prompt and
shared CAD-delivery instructions, with tool access and feedback
determined by the experimental condition described in Appendix~A.

\subsection{Attempts and Time Accounting}
\label{app:time-accounting}

An independent attempt includes construction, testing, and revision
for one mechanism family within a three-hour active-time budget.
In the single-attempt settings, each family receives one such attempt; internal tests,
retries, and revisions do not count as additional independent attempts.

Active time is computed from the union of observed working intervals,
so overlapping activity is counted only once.
For intervals \(I_1,\ldots,I_m\), we define
\begin{equation}
T_{\mathrm{active}}
=
\operatorname{duration}
\left(
\bigcup_{j=1}^{m} I_j
\right).
\end{equation}

Each formal native simulation test has a wall-clock limit of
300 seconds, with simulated duration recorded separately.

\subsection{Artifact Selection and Outcome Aggregation}
\label{app:outcome-aggregation}

Acceptance is determined from the final delivered configuration under
its final declared operating target, following the completion rules
in Appendix~\ref{app:state-completion}.
The record retains the original target, any adjustments, and the
conditions actually evaluated. Each reported outcome is linked to the evaluated design revision,
its test protocol, and the corresponding execution evidence.

Each attempt is assigned to one of three mutually exclusive outcome
groups.
\textit{First-submission pass} denotes completion on the first candidate
submitted to the formal harness.
\textit{Pass after revision} denotes completion on a subsequent candidate
after revision.
\textit{Not passed} denotes the absence of recorded completion by the
end of the attempt, including cases with unfinished evaluation.

Submission order is established from program events and preparation
snapshots rather than revision identifiers alone.
The first formal submission may follow substantial development and
self-testing, so first-submission success does not imply success on
the first generated design.

The three outcome counts sum to the number of available attempts for
each configuration, and total completion is the sum of first-submission
and post-revision passes.
For the single-attempt configurations, Fig.~\ref{fig:model-comparison}
reports median active time across all 46 tasks, including those without
recorded success.

Token use is the median of cumulative constructor input and output tokens per task, in millions, over tasks with recorded usage.
Estimated API cost is computed per task from uncached input, cache reads,
cache writes, and output, then summarized by the median.
We use official standard API prices checked on September 25, 2026,
with short-context rates for Astra and Grok and 5-minute cache writes
for Fable.
The exported counters do not resolve per-request context tiers or cache
lifetimes, so long-context surcharges and 1-hour cache-write rates are
not included.
These estimates cover recorded constructor usage only, excluding
reviewer usage and external tool charges; unreported turns can leave
a task's cost incomplete.

The Grok + history setting allows up to eight sequential attempts per task on the same 46-task cohort, with access to earlier records. Its generated machines undergo the same final batch evaluation as the other settings.

\clearpage
\subsection{Task-Level Completion Results}
\label{app:task-completion}

Table~\ref{tab:task-completion} lists the task-level outcomes for the four single-attempt full-harness configurations underlying
Fig.~\ref{fig:model-comparison}. Task IDs and names follow the
mechanism catalog in Appendix~\ref{app:mechanism-catalog}.

\begingroup
\small
\setlength{\tabcolsep}{4pt}
\renewcommand{\arraystretch}{1.06}
\setlength{\LTleft}{0pt}
\setlength{\LTright}{0pt}
\setlength{\LTcapwidth}{\linewidth}
\captionsetup{singlelinecheck=false}
\begin{longtable}{@{}
    >{\raggedleft\arraybackslash}p{\dimexpr0.055\linewidth-\tabcolsep\relax}
    >{\raggedright\arraybackslash}p{\dimexpr0.475\linewidth-2\tabcolsep\relax}
    *{3}{>{\centering\arraybackslash}p{\dimexpr0.1175\linewidth-2\tabcolsep\relax}}
    >{\centering\arraybackslash}p{\dimexpr0.1175\linewidth-\tabcolsep\relax}
@{}}
\caption{Task-level recorded completion in the four single-attempt full-harness configurations. A check mark denotes completion under formal evaluation; a dash denotes no recorded completion, including unfinished evaluation. Each configuration has one independent attempt per task.}
\label{tab:task-completion}\\
\toprule
\textbf{ID} & \textbf{Mechanism} & \textbf{Astra} & \textbf{Kimi} & \textbf{Fable} & \textbf{Grok} \\
\midrule
\endfirsthead
\multicolumn{6}{@{}l}{\textit{\tablename~\thetable{} continued}}\\[4pt]
\toprule
\textbf{ID} & \textbf{Mechanism} & \textbf{Astra} & \textbf{Kimi} & \textbf{Fable} & \textbf{Grok} \\
\midrule
\endhead
\midrule
\multicolumn{6}{r@{}}{\textit{Continued on next page}}\\
\endfoot
\bottomrule
\endlastfoot
1 & Hinged toolbox & --- & --- & --- & --- \\
2 & Rack-and-pinion gate & \(\checkmark\) & \(\checkmark\) & \(\checkmark\) & \(\checkmark\) \\
3 & Four-bar wiper & \(\checkmark\) & \(\checkmark\) & \(\checkmark\) & \(\checkmark\) \\
4 & Hand-cranked egg beater & \(\checkmark\) & \(\checkmark\) & \(\checkmark\) & \(\checkmark\) \\
5 & Self-locking utility knife & \(\checkmark\) & --- & \(\checkmark\) & \(\checkmark\) \\
6 & Folding drying rack & --- & --- & \(\checkmark\) & --- \\
7 & Bicycle & --- & --- & --- & --- \\
8 & Pedal-operated bin & \(\checkmark\) & --- & \(\checkmark\) & \(\checkmark\) \\
9 & Mechanical alarm clock & --- & --- & --- & --- \\
10 & Kitchen tongs & --- & --- & \(\checkmark\) & --- \\
11 & Retractable ballpoint pen & --- & --- & \(\checkmark\) & --- \\
12 & Nail clipper & \(\checkmark\) & --- & \(\checkmark\) & --- \\
13 & One-way ratchet screwdriver & --- & --- & \(\checkmark\) & --- \\
14 & Linearly actuated scissor lift & --- & --- & \(\checkmark\) & --- \\
15 & Equal-arm balance & \(\checkmark\) & \(\checkmark\) & \(\checkmark\) & \(\checkmark\) \\
16 & Toggle clamp / press & --- & \(\checkmark\) & \(\checkmark\) & \(\checkmark\) \\
17 & Dry piston-pump mechanism & \(\checkmark\) & \(\checkmark\) & \(\checkmark\) & \(\checkmark\) \\
18 & Three-digit combination lock & \(\checkmark\) & --- & \(\checkmark\) & --- \\
19 & Two-digit mechanical counter & \(\checkmark\) & --- & \(\checkmark\) & --- \\
20 & Three-ball dispenser & --- & --- & --- & --- \\
21 & Gravity timer & \(\checkmark\) & --- & \(\checkmark\) & \(\checkmark\) \\
22 & Slider-crank push rod & \(\checkmark\) & \(\checkmark\) & \(\checkmark\) & \(\checkmark\) \\
23 & Gravity-pawl turntable & \(\checkmark\) & --- & --- & --- \\
24 & Sliding-bolt gate & \(\checkmark\) & \(\checkmark\) & \(\checkmark\) & \(\checkmark\) \\
25 & Three-stage gear reducer & --- & --- & \(\checkmark\) & --- \\
26 & Desktop display turntable & \(\checkmark\) & \(\checkmark\) & \(\checkmark\) & \(\checkmark\) \\
27 & Small swing gate & \(\checkmark\) & \(\checkmark\) & \(\checkmark\) & \(\checkmark\) \\
28 & Tilting desktop mirror & \(\checkmark\) & \(\checkmark\) & \(\checkmark\) & \(\checkmark\) \\
29 & Folding wall hook & \(\checkmark\) & \(\checkmark\) & \(\checkmark\) & \(\checkmark\) \\
30 & Spring clothespin & --- & --- & \(\checkmark\) & \(\checkmark\) \\
31 & Scissors & \(\checkmark\) & \(\checkmark\) & \(\checkmark\) & \(\checkmark\) \\
32 & Ratchet hand winch & --- & --- & --- & --- \\
33 & Spring-latch door lock & \(\checkmark\) & --- & --- & \(\checkmark\) \\
34 & Reversible dual-pawl ratchet wrench & --- & --- & --- & --- \\
35 & Full-cycle ratchet crimper & --- & --- & \(\checkmark\) & --- \\
36 & Geneva indexing table with overload clutch & --- & --- & --- & --- \\
37 & Folding shopping cart & \(\checkmark\) & --- & \(\checkmark\) & --- \\
38 & Mechanical \(4\times4\) puzzle cube & --- & --- & --- & --- \\
39 & Twin-cylinder engine linkage & \(\checkmark\) & \(\checkmark\) & \(\checkmark\) & \(\checkmark\) \\
40 & Four-station indexing table & --- & --- & \(\checkmark\) & --- \\
41 & Synchronized parallel gripper & --- & --- & \(\checkmark\) & --- \\
42 & Releasable reduction winch & --- & --- & --- & --- \\
43 & Baby stroller & --- & --- & --- & --- \\
44 & Height- and tilt-adjustable worktable & --- & --- & \(\checkmark\) & --- \\
45 & Coin sorter and counter & --- & --- & --- & --- \\
46 & Interchangeable-tune music box & --- & --- & --- & --- \\
\midrule
\multicolumn{2}{@{}l}{\textbf{Total completed}} &
\textbf{22/46} & \textbf{14/46} & \textbf{32/46} & \textbf{19/46} \\
\end{longtable}
\endgroup

\subsection{Demand-Weighted Completion and Complexity Strata}
\label{app:demand-weighted-completion}

The radar in Fig.~\ref{fig:model-comparison} uses five requirement-based demand grades derived from the taxonomy. Motion coordination grades motion primitives; structural coupling grades topology; functional integration grades coordination among functions and states; contact coordination grades engagement and contact-state requirements; and geometric constraints grade fits, swept clearances, and interacting geometric relations. A grade of 1 denotes a local or direct requirement, 2 denotes a coupled relation, and 3 denotes multiple interacting relations or operating states, as specified below. These are descriptive ordinal annotations, rather than calibrated difficulty units or independently measured cognitive abilities. The functional-integration grade and the original low/medium/high complexity label are related annotations but are not interchangeable.

For each axis, weighted completion is the sum of that axis's grades on completed tasks divided by the sum on all 46 tasks. Using grades 1, 2, and 3 as weights is a linear weighting convention; every task contributes positively to every denominator. The same weights are used for all models.

\begin{table}[!htbp]
\centering
\small
\caption{Anchors for the three mechanical-demand grades.}
\label{tab:demand-grade-rubric}
\begin{tabularx}{\linewidth}{@{}lYYY@{}}
\toprule
Axis & Grade 1 & Grade 2 & Grade 3 \\
\midrule
Motion & Direct operation of one degree of freedom & One input--output relation & Multiple stages or coordinated outputs \\
Structure & Local support or connection & One jointly constrained chain & Multiple constraint groups sharing parts or state \\
Function & One basic operation and return & Coupled functions or operating conditions & Interdependent functions or conditional states \\
Contact & Continuous support or transmission & Engagement/release or force-dependent function & Several contacts coordinated across states \\
Geometry & Local fit, alignment, or retention & Swept clearance or one meshing relation & Interacting clearances, retention, or meshing constraints \\
\bottomrule
\end{tabularx}
\end{table}

\begin{table}[!htbp]
\centering
\small
\setlength{\tabcolsep}{4pt}
\caption{Task demand grades used in the weighted-completion analysis. Columns M, S, F, C, and G denote motion, structure, function, contact, and geometry. Task IDs follow the mechanism catalog.}
\label{tab:task-demand-grades}
\begin{tabular}{@{}rrrrrr@{\qquad}rrrrrr@{}}
\toprule
Task & M & S & F & C & G & Task & M & S & F & C & G \\
\midrule
1 & 1 & 1 & 1 & 1 & 1 & 24 & 2 & 2 & 2 & 2 & 2 \\
2 & 2 & 2 & 2 & 1 & 2 & 25 & 3 & 2 & 2 & 1 & 3 \\
3 & 2 & 2 & 1 & 1 & 2 & 26 & 1 & 1 & 1 & 1 & 1 \\
4 & 3 & 2 & 2 & 1 & 2 & 27 & 1 & 1 & 1 & 1 & 1 \\
5 & 1 & 2 & 3 & 3 & 2 & 28 & 1 & 1 & 1 & 1 & 1 \\
6 & 3 & 3 & 2 & 2 & 3 & 29 & 1 & 1 & 1 & 1 & 1 \\
7 & 3 & 3 & 3 & 2 & 3 & 30 & 2 & 2 & 2 & 2 & 1 \\
8 & 2 & 2 & 2 & 1 & 2 & 31 & 1 & 1 & 1 & 1 & 2 \\
9 & 3 & 3 & 3 & 3 & 3 & 32 & 2 & 3 & 3 & 3 & 2 \\
10 & 2 & 2 & 2 & 2 & 1 & 33 & 2 & 3 & 3 & 3 & 3 \\
11 & 2 & 2 & 3 & 3 & 2 & 34 & 3 & 3 & 3 & 3 & 3 \\
12 & 2 & 2 & 2 & 2 & 2 & 35 & 3 & 3 & 3 & 3 & 3 \\
13 & 2 & 2 & 2 & 2 & 2 & 36 & 3 & 3 & 3 & 3 & 3 \\
14 & 2 & 2 & 2 & 2 & 2 & 37 & 2 & 2 & 2 & 1 & 2 \\
15 & 2 & 2 & 2 & 1 & 1 & 38 & 3 & 3 & 3 & 2 & 3 \\
16 & 2 & 2 & 2 & 2 & 2 & 39 & 3 & 3 & 1 & 1 & 3 \\
17 & 2 & 2 & 2 & 1 & 2 & 40 & 2 & 2 & 2 & 2 & 2 \\
18 & 3 & 3 & 3 & 3 & 3 & 41 & 3 & 2 & 2 & 2 & 2 \\
19 & 3 & 3 & 3 & 3 & 3 & 42 & 2 & 3 & 3 & 3 & 2 \\
20 & 2 & 3 & 3 & 3 & 3 & 43 & 3 & 3 & 2 & 2 & 2 \\
21 & 2 & 2 & 3 & 2 & 2 & 44 & 3 & 3 & 3 & 2 & 2 \\
22 & 2 & 2 & 2 & 1 & 2 & 45 & 3 & 3 & 3 & 3 & 3 \\
23 & 2 & 2 & 2 & 2 & 2 & 46 & 3 & 3 & 3 & 3 & 3 \\
\bottomrule
\end{tabular}
\end{table}

\begin{table}[!htbp]
\centering
\small
\caption{Recorded completion within the benchmark's original complexity classes. These classes were not constructed by binning the radar scores.}
\label{tab:complexity-completion}
\begin{tabular}{@{}lrrrrr@{}}
\toprule
Complexity & Tasks & Astra & Kimi & Fable & Grok \\
\midrule
Low & 11 & 9/11 & 8/11 & 9/11 & 9/11 \\
Medium & 20 & 12/20 & 6/20 & 18/20 & 10/20 \\
High & 15 & 1/15 & 0/15 & 5/15 & 0/15 \\
\bottomrule
\end{tabular}
\end{table}

\subsection{Tool-Use Annotation and Execution-Time Profiles}
\label{app:tool-use-analysis}

\paragraph{Tool-call classification.}
Each deduplicated constructor tool call receives one primary-purpose
label based on its arguments, outputs, and any recoverable helper-script
content from preceding events.
\textit{Information access} covers reading task requirements, tool
documentation, and source code.
\textit{Design, build, and test implementation} covers source edits,
artifact construction, and test writing.
\textit{Geometry checks} include explicit validation calls, custom
geometric measurements, and motion-geometry audits.
\textit{Simulation} includes development runs, cost probes, and custom
dynamics trials.
\textit{Visual evidence} includes preview generation and image-access
operations, while \textit{feedback analysis} covers reading test results
and analyzing recorded traces.
\textit{Execution and workflow management} includes environment setup,
file transfer, process monitoring, and artifact bookkeeping.
Mixed-purpose calls and calls whose purpose cannot be resolved are
assigned to \textit{Others}.

These labels describe observed actions rather than their success or
the model's underlying reasoning ability.
In particular, an image-access operation does not establish that the
model successfully inspected or understood the image.

\paragraph{Sample selection and normalization.}
The analysis uses 46 task families with observed standard-tool execution
time for all four configurations.
Calls with missing or truncated arguments that cannot be recovered are
excluded before normalization.
For this matched sample, 1,883 Kimi calls and one Grok call are excluded,
leaving 6,035, 6,018, 8,474, and 22,223 calls for Astra, Kimi, Fable, and
Grok, respectively.

Category shares are first computed within each task and then averaged
with equal weight across tasks.
This procedure prevents tasks with unusually many tool calls from
dominating the aggregate profile.

\paragraph{Execution-time accounting.}
Execution-time shares are computed from independently recorded start
and end times for standard build, check, simulate, probe, and preview
operations.
Overlapping intervals within the same category are counted once, while
intervals containing simultaneous activity from multiple categories are
assigned to \textit{Concurrent}.
Intervals with missing endpoints or unresolved zero duration are
excluded.

For each task, category durations are normalized by the total observed
standard-tool execution time and then averaged equally across the same 41 families.
These profiles include preparation performed within each tool operation,
but do not represent total agent runtime or all categories in the
behavioral analysis.
Preview time measures rendering, and neither it nor the remaining
portion of active time is interpreted as a direct measure of model
reasoning.

\paragraph{Timing-coverage check.}
We additionally examine the 14 task families with complete standard-tool
timing records for all four configurations.
Simulation remains the largest execution-time category for every
configuration, and Grok retains the highest share of build time.
These comparisons describe observed execution costs, which depend on
the generated mechanisms and execution environments as well as the
agents' tool choices.

\section{Detailed Harness Comparisons}
\label{app:posthoc-evaluation}

Table~\ref{tab:harness-overlaps} retains the success-set overlaps behind Table~\ref{tab:harness-ablations}. Independent agents evaluate the final assemblies from each workflow under the task requirements and fixed constraints, selecting simulation strategies and configuring candidate-specific tests as described in Section~\ref{sec:model-comparison}.

\begin{table}[htbp]
\centering
\small
\setlength{\tabcolsep}{5pt}
\caption{Whole-task successes and their overlap with the full harness. Shared counts tasks solved by both the row's workflow and the full harness for the same model; Only counts successes unique to that workflow relative to the full harness.}
\label{tab:harness-overlaps}
\begin{tabular}{@{}lrrrrrr@{}}
\toprule
& \multicolumn{3}{c}{Astra} & \multicolumn{3}{c}{Grok} \\
\cmidrule(lr){2-4}\cmidrule(l){5-7}
Setting & Passed & Shared & Only & Passed & Shared & Only \\
\midrule
Requirements only & 1/46 & 1 & 0 & 3/46 & 3 & 0 \\
Simulation tools & 8/46 & 7 & 1 & 10/46 & 8 & 2 \\
Full harness & 22/46 & --- & --- & 19/46 & --- & --- \\
Binary feedback & 0/46 & 0 & 0 & 8/46 & 7 & 1 \\
\bottomrule
\end{tabular}
\end{table}

Relative to simulation tools, the full harness has 15 additional successful tasks for Astra and 11 for Grok; the tools-only setting has one and two successes outside the respective full-harness sets. The net increases are therefore 14 and nine tasks. Relative to simulation tools, both full-harness configurations additionally solve the rack-and-pinion gate, four-bar wiper, and hand-cranked whisk (Tasks~2--4), as well as the self-locking knife and spring-latch lock (Tasks~5 and~33). In the binary-feedback setting, Grok has seven successes shared with the full harness and one additional success.

\clearpage
The comparisons in Section~\ref{sec:harness-ablations} examine three roles of the harness: providing access to physical simulation, maintaining coverage of the functional requirements, and returning evidence that guides revision. This appendix specifies the information available in each setting, lists the task-level outcomes, and uses selected trajectories to examine how agents test, revise, and decide when to stop.

\subsection{Experimental Settings and Evaluation}
\label{app:ablation-protocol}

\paragraph{Construction-time support.}
All four settings start from the task's functional requirements and permit CAD construction, source inspection, and static visualization. Table~\ref{tab:ablation-support} distinguishes the supplied simulation support, responsibility for test coverage, and visible feedback. In the \emph{requirements-only} setting, agents may still write verification code, including their own simulations. The \emph{simulation-tools} setting supplies a simulator and guidance, while leaving the agent to select operating conditions, observations, and acceptance checks. The \emph{full harness} additionally connects the candidate mechanism to requirement-level tests throughout design revision.

\begin{table}[!htbp]
\centering
\small
\arrayrulecolor{black}
\setlength{\tabcolsep}{5pt}
\renewcommand{\arraystretch}{1.15}
\caption{Support and feedback available during construction.}
\label{tab:ablation-support}
\begin{tabularx}{\linewidth}{@{}p{0.18\linewidth}YYY@{}}
\toprule
\textbf{Setting} & \textbf{Simulation support} & \textbf{Requirement-level testing} & \textbf{Visible evidence} \\
\midrule
Requirements only & Agent implements any simulation it chooses to use. & Agent chooses its own checks. & Outputs of agent-written checks. \\
\addlinespace[3pt]
Simulation tools & Simulator, starter code, and tool guidance are supplied. & Agent chooses its own checks. & Agent's simulation outputs and geometric analyses. \\
\addlinespace[3pt]
Binary feedback & Harness constructs and executes candidate-specific tests. & Required functions remain part of the evaluation workflow. & Pass/fail outcomes; execution and validation diagnostics are hidden. \\
\addlinespace[3pt]
Full harness & Harness constructs and executes candidate-specific tests. & Required functions remain part of the evaluation workflow. & Failed checks, measured discrepancies, and physical-validity diagnostics. \\
\bottomrule
\end{tabularx}
\end{table}

\paragraph{Scope of binary feedback.}
The binary-feedback setting restricts both development and formal simulation tests, as well as the standard validation interface, to pass/fail outcomes. The constructor cannot inspect the execution traces, measured residuals, validator diagnostics, or reviewer artifacts underlying these outcomes. It can still inspect its CAD source and static previews, perform its own geometric or analytical calculations, and request additional tests. This comparison asks whether correctness signals alone provide sufficient direction for repair. Simulation tools and binary feedback change both test organization and evidence access; they are complementary comparisons rather than successive levels of a single support variable.

\paragraph{Evaluation and outcome accounting.}
Requirements-only and simulation-tools designs are adapted for evaluation after construction by a separate agent, whereas the harness settings expose requirement-level evaluation during revision. Post-hoc evaluation includes establishing an executable physical model of the delivered assembly. An incomplete adaptation or execution therefore leaves whole-task completion unestablished, without by itself identifying a mechanical defect. We report the same recorded whole-task outcomes as Table~\ref{tab:harness-ablations}; passing an agent-written check is evidence only for that check's scope. The binary-feedback analysis uses the isolated 46-task batch for each model, excluding the earlier development and pilot batches.

\clearpage
\subsection{Task-Level Outcomes}
\label{app:ablation-task-outcomes}

Table~\ref{tab:ablation-task-outcomes} expands Table~\ref{tab:harness-ablations} by task. Full-harness outcomes reuse the main-experiment runs; each other setting contributes one selected run per model and task.

\begin{table}[!htbp]
\centering
\small
\arrayrulecolor{black}
\setlength{\tabcolsep}{3pt}
\renewcommand{\arraystretch}{1.0}
\captionsetup{singlelinecheck=false}
\caption{Recorded completion by task and setting. Req.: requirements only; Sim.: simulation tools; Bin.: binary feedback. A check mark denotes recorded whole-task completion; a dash denotes no recorded completion, including incomplete evaluation or partial delivery.}
\label{tab:ablation-task-outcomes}
\begin{tabular}{@{}
    >{\raggedleft\arraybackslash}p{\dimexpr0.045\linewidth-\tabcolsep\relax}
    >{\raggedright\arraybackslash}p{\dimexpr0.435\linewidth-2\tabcolsep\relax}
    *{7}{>{\centering\arraybackslash}p{\dimexpr0.065\linewidth-2\tabcolsep\relax}}
    >{\centering\arraybackslash}p{\dimexpr0.065\linewidth-\tabcolsep\relax}
@{}}
\toprule
 & & \multicolumn{4}{c}{\textbf{Astra}} & \multicolumn{4}{c}{\textbf{Grok}} \\
\cmidrule(lr){3-6}\cmidrule(l){7-10}
\textbf{ID} & \textbf{Mechanism} & Req. & Sim. & Bin. & Full & Req. & Sim. & Bin. & Full \\
\midrule

1 & Hinged toolbox & --- & --- & --- & --- & --- & --- & --- & --- \\
2 & Rack-and-pinion gate & --- & --- & --- & \(\checkmark\) & --- & --- & --- & \(\checkmark\) \\
3 & Four-bar wiper & --- & --- & --- & \(\checkmark\) & --- & --- & --- & \(\checkmark\) \\
4 & Hand-cranked egg beater & --- & --- & --- & \(\checkmark\) & --- & --- & --- & \(\checkmark\) \\
5 & Self-locking utility knife & --- & --- & --- & \(\checkmark\) & --- & --- & --- & \(\checkmark\) \\
6 & Folding drying rack & --- & --- & --- & --- & --- & --- & --- & --- \\
7 & Bicycle & --- & --- & --- & --- & --- & --- & --- & --- \\
8 & Pedal-operated bin & --- & \(\checkmark\) & --- & \(\checkmark\) & --- & \(\checkmark\) & --- & \(\checkmark\) \\
9 & Mechanical alarm clock & --- & --- & --- & --- & --- & --- & --- & --- \\
10 & Kitchen tongs & --- & --- & --- & --- & --- & \(\checkmark\) & --- & --- \\
11 & Retractable ballpoint pen & --- & --- & --- & --- & --- & --- & --- & --- \\
12 & Nail clipper & --- & --- & --- & \(\checkmark\) & --- & --- & \(\checkmark\) & --- \\
13 & One-way ratchet screwdriver & --- & --- & --- & --- & --- & --- & --- & --- \\
14 & Linearly actuated scissor lift & --- & --- & --- & --- & --- & --- & --- & --- \\
15 & Equal-arm balance & --- & --- & --- & \(\checkmark\) & --- & --- & \(\checkmark\) & \(\checkmark\) \\
16 & Toggle clamp / press & --- & \(\checkmark\) & --- & --- & --- & \(\checkmark\) & --- & \(\checkmark\) \\
17 & Dry piston-pump mechanism & --- & \(\checkmark\) & --- & \(\checkmark\) & --- & \(\checkmark\) & --- & \(\checkmark\) \\
18 & Three-digit combination lock & --- & --- & --- & \(\checkmark\) & --- & --- & --- & --- \\
19 & Two-digit mechanical counter & --- & --- & --- & \(\checkmark\) & --- & --- & --- & --- \\
20 & Three-ball dispenser & --- & --- & --- & --- & --- & --- & --- & --- \\
21 & Gravity timer & --- & --- & --- & \(\checkmark\) & --- & --- & --- & \(\checkmark\) \\
22 & Slider-crank push rod & --- & --- & --- & \(\checkmark\) & --- & --- & \(\checkmark\) & \(\checkmark\) \\
23 & Gravity-pawl turntable & --- & \(\checkmark\) & --- & \(\checkmark\) & --- & \(\checkmark\) & --- & --- \\
24 & Sliding-bolt gate & --- & --- & --- & \(\checkmark\) & --- & --- & \(\checkmark\) & \(\checkmark\) \\
25 & Three-stage gear reducer & --- & --- & --- & --- & --- & --- & --- & --- \\
26 & Desktop display turntable & \(\checkmark\) & --- & --- & \(\checkmark\) & --- & \(\checkmark\) & \(\checkmark\) & \(\checkmark\) \\
27 & Small swing gate & --- & \(\checkmark\) & --- & \(\checkmark\) & \(\checkmark\) & \(\checkmark\) & \(\checkmark\) & \(\checkmark\) \\
28 & Tilting desktop mirror & --- & --- & --- & \(\checkmark\) & \(\checkmark\) & --- & \(\checkmark\) & \(\checkmark\) \\
29 & Folding wall hook & --- & \(\checkmark\) & --- & \(\checkmark\) & --- & \(\checkmark\) & \(\checkmark\) & \(\checkmark\) \\
30 & Spring clothespin & --- & --- & --- & --- & --- & --- & --- & \(\checkmark\) \\
31 & Scissors & --- & \(\checkmark\) & --- & \(\checkmark\) & --- & \(\checkmark\) & --- & \(\checkmark\) \\
32 & Ratchet hand winch & --- & --- & --- & --- & --- & --- & --- & --- \\
33 & Spring-latch door lock & --- & --- & --- & \(\checkmark\) & --- & --- & --- & \(\checkmark\) \\
34 & Reversible dual-pawl ratchet wrench & --- & --- & --- & --- & --- & --- & --- & --- \\
35 & Full-cycle ratchet crimper & --- & --- & --- & --- & --- & --- & --- & --- \\
36 & Geneva indexing table with overload clutch & --- & --- & --- & --- & --- & --- & --- & --- \\
37 & Folding shopping cart & --- & --- & --- & \(\checkmark\) & --- & --- & --- & --- \\
38 & Mechanical \(4\times4\) puzzle cube & --- & --- & --- & --- & --- & --- & --- & --- \\
39 & Twin-cylinder engine linkage & --- & \(\checkmark\) & --- & \(\checkmark\) & \(\checkmark\) & \(\checkmark\) & --- & \(\checkmark\) \\
40 & Four-station indexing table & --- & --- & --- & --- & --- & --- & --- & --- \\
41 & Synchronized parallel gripper & --- & --- & --- & --- & --- & --- & --- & --- \\
42 & Releasable reduction winch & --- & --- & --- & --- & --- & --- & --- & --- \\
43 & Baby stroller & --- & --- & --- & --- & --- & --- & --- & --- \\
44 & Height- and tilt-adjustable worktable & --- & --- & --- & --- & --- & --- & --- & --- \\
45 & Coin sorter and counter & --- & --- & --- & --- & --- & --- & --- & --- \\
46 & Interchangeable-tune music box & --- & --- & --- & --- & --- & --- & --- & --- \\
\midrule
\multicolumn{2}{@{}l}{\textbf{Total completed}} & \textbf{1} & \textbf{8} & \textbf{0} & \textbf{22} & \textbf{3} & \textbf{10} & \textbf{8} & \textbf{19} \\
\bottomrule
\end{tabular}
\end{table}

\clearpage
\subsection{Verification Scope and Revision Without the Harness}
\label{app:ablation-verification}

\paragraph{Access to simulation supports meaningful engineering.}
The simulation-tools trajectories contain active physical diagnosis and geometric repair. For example, Astra adds a detent to an indexing table after identifying motion during the intended dwell phase, then checks recovery from obstructions. Such behavior shows that agents can formulate useful experiments without the harness. The question is how those experiments collectively establish the requested function, and whether that target remains intact as the design and test code change.

\begin{table}[!htbp]
\centering
\small
\arrayrulecolor{black}
\setlength{\tabcolsep}{5pt}
\renewcommand{\arraystretch}{1.12}
\caption{Testing and revision in the simulation-tools setting. Agent-written checks are distinguished from recorded whole-task completion.}
\label{tab:ablation-verification-cases}
\begin{tabularx}{\linewidth}{@{}>{\raggedright\arraybackslash}p{0.135\linewidth}YYY@{}}
\toprule
\textbf{Case} & \textbf{Agent action} & \textbf{Observed evidence} & \textbf{Implication and remaining scope} \\
\midrule
\textbf{Task 36}\newline Astra\newline Geneva indexing table &
Adds a spring-loaded roller detent after station drift; tests obstructions near entry and exit; relieves slot mouths after a swept-geometry check. &
The constructor reports eight $90^{\circ}$ steps, dwell holding under disturbances, and recovery after obstruction in its own simulations. &
Task-relevant experiments lead to concrete mechanical changes. These local results do not establish whole-task completion in the reported evaluation. \\
\addlinespace[6pt]
\textbf{Task 9}\newline Astra\newline Mechanical alarm clock &
Tests the escapement for 600 simulated seconds and tests the striker separately; revises gear phasing and local clearances. &
The escapement advances approximately ten revolutions. The delivery explicitly leaves the integrated alarm sequence unverified; the hour wheel also interferes with the alarm lever. &
Subsystem evidence leaves a composition gap: setting, triggering, and stopping the alarm must all preserve timekeeping. See Fig.~\ref{fig:astra-functional-failures}b. \\
\addlinespace[6pt]
\textbf{Task 13}\newline Grok\newline Ratchet screwdriver &
Replaces a numerically invalid 1.40-second loaded, two-stroke test with a 0.20-second stepping test. &
The new run returns 21 finite samples, but the bit remains nearly stationary. The original drive and return-stroke checks are not repeated. &
Restoring executable simulation narrows the tested behavior. The agent identifies the result as a smoke test and does not claim functional success. \\
\bottomrule
\end{tabularx}
\end{table}

\paragraph{Complete functions span interactions and operation sequences.}
The clock case illustrates why successful component tests do not establish the behavior of an assembled mechanism. The missing evidence concerns interactions between subsystems and the preservation of timekeeping through an alarm sequence. A requirement-level testbed makes these obligations explicit alongside tests of the escapement and striker. This connects functional decomposition to composition: individual tests must jointly cover the user's requested behavior, including transitions between operations.

\paragraph{A revision should preserve the target being tested.}
The screwdriver case separates numerical recovery from functional recovery. Producing finite simulation states is useful, but it leaves unanswered whether the handle advances the output under load without undoing that motion on return. In a separate full-harness Grok run on Task~13, the second candidate passes both forward-advance and both handle-return checks, while both output-backslip checks remain failures. Although neither run records whole-task completion, the harness result identifies the outstanding requirement after partial progress. These are independently generated candidates; the comparison illustrates how the evaluation target is maintained during revision.

\clearpage
\subsection{Diagnosis and Stopping Under Binary Feedback}
\label{app:ablation-binary-diagnosis}

\paragraph{Binary outcomes shift diagnosis toward additional experiments.}
A failed test can reflect an incorrect input convention, obstructed motion, or an invalid execution setup. A pass/fail response does not distinguish these explanations. Table~\ref{tab:ablation-binary-cases} shows three responses: designing tests that separate hypotheses, reconstructing evidence through independent geometric reasoning, and revising several plausible failure sources without confirming which caused the rejection.

\begin{table}[!htbp]
\centering
\small
\arrayrulecolor{black}
\setlength{\tabcolsep}{5pt}
\renewcommand{\arraystretch}{1.12}
\caption{Diagnostic strategies under binary feedback. Geometric and kinematic calculations are the agents' own analyses, not hidden simulation traces.}
\label{tab:ablation-binary-cases}
\begin{tabularx}{\linewidth}{@{}>{\raggedright\arraybackslash}p{0.135\linewidth}YYY@{}}
\toprule
\textbf{Case} & \textbf{Agent action} & \textbf{Evidence and outcome} & \textbf{What the case shows} \\
\midrule
\textbf{Task 1}\newline Astra\newline Hinged toolbox &
Tests rest and opening with opposite torque signs. Reverses the hinge's body ordering after only the reverse-direction opening test passes. &
A subsequent positive-torque opening test passes. The complete two-cycle operation remains unverified; the agent delivers partial results. &
Carefully chosen binary tests can distinguish an input-direction error from other explanations. Resolving one local cause does not establish the full operating sequence. \\
\addlinespace[6pt]
\textbf{Task 12}\newline Grok\newline Nail clipper &
Uses its own geometry and linkage calculations to investigate input direction, excessive follower travel, and changes in the four-bar solution branch; revises the mechanism. &
The revised design obtains a recorded whole-task pass. Several coupled design changes precede this result. &
Restricted feedback still permits successful diagnosis when independent analysis supplies the missing information. This success cannot be attributed to a single parameter change. \\
\addlinespace[6pt]
\textbf{Task 33}\newline Grok\newline Spring-latch door lock &
Measures clearances and revises hinges, handle bearings, cam orientation, latch geometry, and the strike plate after repeated binary failures. &
The agent finds geometric problems, but the returned failures do not confirm which defect caused rejection. No whole-task completion is recorded. &
Useful inspection can coexist with unresolved failure localization. The full-harness repair in Fig.~\ref{fig:task33-relocking-repair} instead tracks relocking checks and then residual interference. \\
\bottomrule
\end{tabularx}
\end{table}

\paragraph{The models also differ in when they stop.}
Table~\ref{tab:ablation-binary-stopping} summarizes the recorded termination states in the binary-feedback batch. Astra chooses partial delivery in 43 of 46 runs, with a median of 137.8 minutes of active budget remaining at those decisions. Its explanations repeatedly cite insufficient evidence for a targeted next revision. Grok instead reaches the budget limit in 36 runs and completes eight. The same restricted feedback thus accompanies different search and stopping strategies; Astra's zero recorded completions should not be read as 46 fully executed mechanical failures.

\begin{table}[!htbp]
\centering
\small
\arrayrulecolor{black}
\setlength{\tabcolsep}{6pt}
\renewcommand{\arraystretch}{1.1}
\caption{Recorded termination states under binary feedback, with 46 runs per model. Categories are mutually exclusive. Evidence required denotes a recorded unresolved evidence state.}
\label{tab:ablation-binary-stopping}
\begin{tabularx}{\linewidth}{@{}l*{4}{>{\centering\arraybackslash}X}@{}}
\toprule
\textbf{Model} & \textbf{Completed} & \textbf{Partial delivery} & \textbf{Budget exhausted} & \textbf{Evidence required} \\
\midrule
Astra & 0 & 43 & 0 & 3 \\
Grok & 8 & 0 & 36 & 2 \\
\bottomrule
\end{tabularx}
\end{table}

Together, these observations separate a correctness signal from evidence for choosing a revision. Binary outcomes can guide repair through additional experiments or independent analysis. Detailed feedback supplies further information about the failed operation or validity condition, helping the agent select and verify its next change. The stopping counts compare models within the binary setting; estimating how masking changes each model's stopping policy would require a paired analysis of the full-harness runs.

\subsection{Construction Without Visual Access}
\label{app:no-vision}

Section~\ref{sec:discussions} notes that agents rarely inspect rendered images during functional design. To test whether visual access nonetheless contributes to task success, we repeat the single-attempt Astra and Grok full-harness experiments with one additional restriction on the constructor. Task prompts, constructor instructions, the harness workflow, physical simulation, diagnostic numerical feedback, the active-time budget, and the independent final evaluation are identical to Section~\ref{sec:model-comparison}. The constructor, however, cannot view, generate, or capture images, and cannot operate a computer or browser through screenshots. It cannot render the model through the harness preview tool, through its own scripts using plotting, imaging, or CAD visualization libraries, or through external modeling and rendering programs. It also cannot open image, video, 3D-viewer, HTML, or PDF files, and harness feedback returned to it contains only text and numerical values. The constructor retains its CAD source and exported geometry (STEP and STL), programmatic geometric queries, and simulation. The test-binding agent, independent review, and final evaluation keep their standard visual access, so the restriction removes visual information only from design and revision.

Table~\ref{tab:no-vision} compares the two runs. Without visual access, Astra completes 20 of the 46 tasks, compared with 22 with visual access. Sixteen tasks are solved in both settings. Six are solved only with visual access: the four-bar wiper, piston pump, two-digit counter, gravity-pawl turntable, tilting mirror, and folding shopping cart (Tasks~3, 17, 19, 23, 28, and~37). Four are solved only without it: the hinged toolbox, scissor lift, toggle clamp, and parallel gripper (Tasks~1, 14, 16, and~41). Grok shows a similar pattern: without visual access, it completes 18 of the 46 tasks, compared with 19 with visual access; 14 tasks are solved in both settings, five only with visual access, and four only without it.

\begin{figure}[!htbp]
\centering
\includegraphics[width=\linewidth]{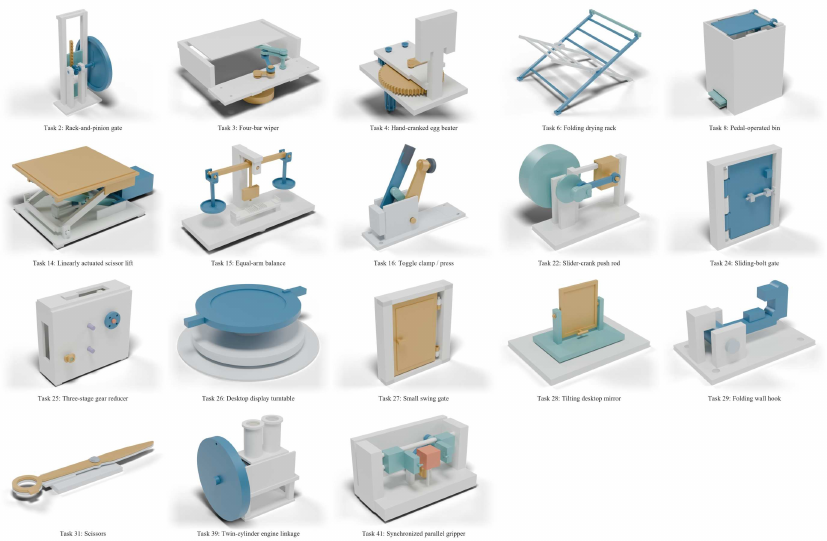}
\caption{\textbf{Mechanisms completed by Grok without visual access for the constructor.}}
\label{fig:no-vision-grok-gallery}
\end{figure}

Removing visual access therefore changes the aggregate count by two tasks, while the sets of solved tasks differ in ten. With one attempt per task, these task-level differences cannot be attributed to visual access alone. The aggregate result nevertheless indicates that, within \harnessname, the constructor's functional success depends mainly on CAD programs, geometric queries, and simulation feedback rather than on visual inspection, consistent with the tool-use profiles in Appendix~\ref{app:tool-use-analysis}. Figure~\ref{fig:no-vision-gallery} shows the 20 designs completed without visual access, and Figure~\ref{fig:no-vision-grok-gallery} shows the 18 designs completed by Grok without visual access.

\begin{figure}[!htbp]
\centering
\includegraphics[width=\linewidth]{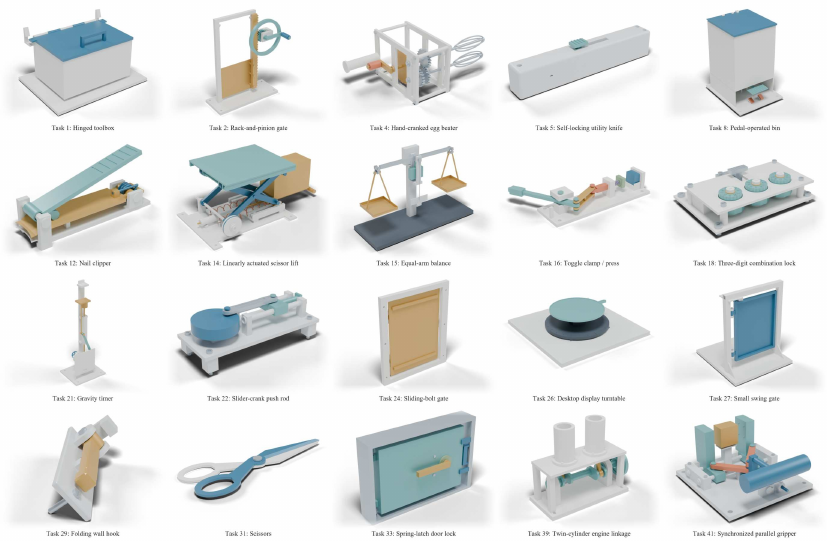}
\caption{\textbf{Mechanisms completed by Astra without visual access for the constructor.}}
\label{fig:no-vision-gallery}
\end{figure}

\begingroup
\small
\setlength{\tabcolsep}{4pt}
\renewcommand{\arraystretch}{1.06}
\setlength{\LTleft}{0pt}
\setlength{\LTright}{0pt}
\setlength{\LTcapwidth}{\linewidth}
\captionsetup{singlelinecheck=false}
\begin{longtable}{@{}
    >{\raggedleft\arraybackslash}p{\dimexpr0.055\linewidth-\tabcolsep\relax}
    >{\raggedright\arraybackslash}p{\dimexpr0.465\linewidth-2\tabcolsep\relax}
    *{3}{>{\centering\arraybackslash}p{\dimexpr0.12\linewidth-2\tabcolsep\relax}}
    >{\centering\arraybackslash}p{\dimexpr0.12\linewidth-\tabcolsep\relax}
@{}}
\caption{Task-level recorded completion of single-attempt Astra and Grok full-harness runs with and without visual access for the constructor. A check mark denotes completion under formal evaluation; a dash denotes no recorded completion, including unfinished evaluation. Each setting has one independent attempt per task; the \emph{With vision} columns reproduce the Astra and Grok columns of Table~\ref{tab:task-completion}.}
\label{tab:no-vision}\\
\toprule
& & \multicolumn{2}{c}{\textbf{Astra}} & \multicolumn{2}{c@{}}{\textbf{Grok}} \\
\cmidrule(lr){3-4}\cmidrule(l){5-6}
\textbf{ID} & \textbf{Mechanism} & \textbf{With vision} & \textbf{No vision} & \textbf{With vision} & \textbf{No vision} \\
\midrule
\endfirsthead
\multicolumn{6}{@{}l}{\textit{\tablename~\thetable{} continued}}\\[4pt]
\toprule
& & \multicolumn{2}{c}{\textbf{Astra}} & \multicolumn{2}{c@{}}{\textbf{Grok}} \\
\cmidrule(lr){3-4}\cmidrule(l){5-6}
\textbf{ID} & \textbf{Mechanism} & \textbf{With vision} & \textbf{No vision} & \textbf{With vision} & \textbf{No vision} \\
\midrule
\endhead
\midrule
\multicolumn{6}{r@{}}{\textit{Continued on next page}}\\
\endfoot
\bottomrule
\endlastfoot
1 & Hinged toolbox & --- & \(\checkmark\) & --- & --- \\
2 & Rack-and-pinion gate & \(\checkmark\) & \(\checkmark\) & \(\checkmark\) & \(\checkmark\) \\
3 & Four-bar wiper & \(\checkmark\) & --- & \(\checkmark\) & \(\checkmark\) \\
4 & Hand-cranked egg beater & \(\checkmark\) & \(\checkmark\) & \(\checkmark\) & \(\checkmark\) \\
5 & Self-locking utility knife & \(\checkmark\) & \(\checkmark\) & \(\checkmark\) & --- \\
6 & Folding drying rack & --- & --- & --- & \(\checkmark\) \\
7 & Bicycle & --- & --- & --- & --- \\
8 & Pedal-operated bin & \(\checkmark\) & \(\checkmark\) & \(\checkmark\) & \(\checkmark\) \\
9 & Mechanical alarm clock & --- & --- & --- & --- \\
10 & Kitchen tongs & --- & --- & --- & --- \\
11 & Retractable ballpoint pen & --- & --- & --- & --- \\
12 & Nail clipper & \(\checkmark\) & \(\checkmark\) & --- & --- \\
13 & One-way ratchet screwdriver & --- & --- & --- & --- \\
14 & Linearly actuated scissor lift & --- & \(\checkmark\) & --- & \(\checkmark\) \\
15 & Equal-arm balance & \(\checkmark\) & \(\checkmark\) & \(\checkmark\) & \(\checkmark\) \\
16 & Toggle clamp / press & --- & \(\checkmark\) & \(\checkmark\) & \(\checkmark\) \\
17 & Dry piston-pump mechanism & \(\checkmark\) & --- & \(\checkmark\) & --- \\
18 & Three-digit combination lock & \(\checkmark\) & \(\checkmark\) & --- & --- \\
19 & Two-digit mechanical counter & \(\checkmark\) & --- & --- & --- \\
20 & Three-ball dispenser & --- & --- & --- & --- \\
21 & Gravity timer & \(\checkmark\) & \(\checkmark\) & \(\checkmark\) & --- \\
22 & Slider-crank push rod & \(\checkmark\) & \(\checkmark\) & \(\checkmark\) & \(\checkmark\) \\
23 & Gravity-pawl turntable & \(\checkmark\) & --- & --- & --- \\
24 & Sliding-bolt gate & \(\checkmark\) & \(\checkmark\) & \(\checkmark\) & \(\checkmark\) \\
25 & Three-stage gear reducer & --- & --- & --- & \(\checkmark\) \\
26 & Desktop display turntable & \(\checkmark\) & \(\checkmark\) & \(\checkmark\) & \(\checkmark\) \\
27 & Small swing gate & \(\checkmark\) & \(\checkmark\) & \(\checkmark\) & \(\checkmark\) \\
28 & Tilting desktop mirror & \(\checkmark\) & --- & \(\checkmark\) & \(\checkmark\) \\
29 & Folding wall hook & \(\checkmark\) & \(\checkmark\) & \(\checkmark\) & \(\checkmark\) \\
30 & Spring clothespin & --- & --- & \(\checkmark\) & --- \\
31 & Scissors & \(\checkmark\) & \(\checkmark\) & \(\checkmark\) & \(\checkmark\) \\
32 & Ratchet hand winch & --- & --- & --- & --- \\
33 & Spring-latch door lock & \(\checkmark\) & \(\checkmark\) & \(\checkmark\) & --- \\
34 & Reversible dual-pawl ratchet wrench & --- & --- & --- & --- \\
35 & Full-cycle ratchet crimper & --- & --- & --- & --- \\
36 & Geneva indexing table with overload clutch & --- & --- & --- & --- \\
37 & Folding shopping cart & \(\checkmark\) & --- & --- & --- \\
38 & Mechanical \(4\times4\) puzzle cube & --- & --- & --- & --- \\
39 & Twin-cylinder engine linkage & \(\checkmark\) & \(\checkmark\) & \(\checkmark\) & \(\checkmark\) \\
40 & Four-station indexing table & --- & --- & --- & --- \\
41 & Synchronized parallel gripper & --- & \(\checkmark\) & --- & \(\checkmark\) \\
42 & Releasable reduction winch & --- & --- & --- & --- \\
43 & Baby stroller & --- & --- & --- & --- \\
44 & Height- and tilt-adjustable worktable & --- & --- & --- & --- \\
45 & Coin sorter and counter & --- & --- & --- & --- \\
46 & Interchangeable-tune music box & --- & --- & --- & --- \\
\midrule
\multicolumn{2}{@{}l}{\textbf{Total completed}} &
\textbf{22/46} & \textbf{20/46} & \textbf{19/46} & \textbf{18/46} \\
\end{longtable}
\endgroup

\subsection{Repeated Attempts Without Shared Records}
\label{app:independent-attempts}

The Grok + history setting in Section~\ref{sec:model-comparison} allows up to eight attempts per task, each with access to records from earlier attempts, and completes 36 of the 46 tasks. To separate the contribution of these records from that of additional sampling, we repeat the eight-attempt Grok experiment with one change: each attempt can access only its own artifacts and cannot read the designs, tests, feedback, or other records produced by any other attempt. Task prompts, constructor instructions, the full-harness workflow, the per-attempt budget, and the final evaluation are otherwise identical. All eight attempts are executed for every task, and a task counts as completed if at least one attempt passes the formal evaluation.

Table~\ref{tab:independent-attempts} lists the outcomes. Individual batches complete between 15 and 18 of the 46 tasks (129 passing runs out of 368, or 16.1 tasks per batch on average), comparable to the single-attempt Grok configuration in Table~\ref{tab:task-completion} (19/46). Taking the union over eight independent attempts raises completion to 24 of the 46 tasks, compared with 36 when attempts share records. Relative to a single attempt, additional independent sampling thus adds five tasks, whereas eight attempts with shared records add 17. The union includes seven tasks not completed by the single-attempt run (Tasks~1, 10, 12, 18, 23, 25, and~38) but misses two that it completes, the self-locking utility knife and gravity timer (Tasks~5 and~21), reflecting run-to-run variation at a fixed configuration.

\begin{figure}[!htbp]
\centering
\includegraphics[width=\linewidth]{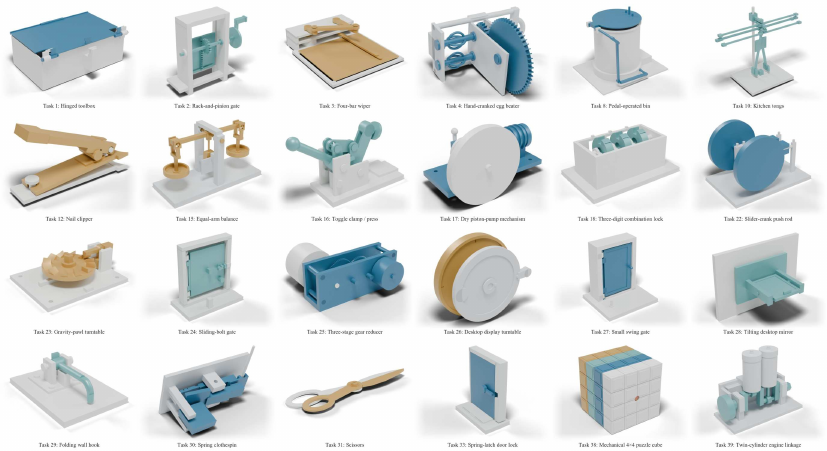}
\caption{\textbf{Mechanisms completed by Grok in pass@8 without shared records.}}
\label{fig:independent-attempts-gallery}
\end{figure}

Outcomes across attempts are also uneven. Seven tasks pass in all eight attempts, whereas six of the 24 completed tasks pass in at most two, and 22 tasks are completed by no attempt. Without shared records, each attempt restarts from the task description, so repeated sampling mainly recovers tasks that a single attempt already solves with moderate probability. Figure~\ref{fig:independent-attempts-gallery} shows one passing design for each of the 24 tasks completed by at least one independent attempt.

\begingroup
\small
\setlength{\tabcolsep}{4pt}
\renewcommand{\arraystretch}{1.06}
\setlength{\LTleft}{0pt}
\setlength{\LTright}{0pt}
\setlength{\LTcapwidth}{\linewidth}
\captionsetup{singlelinecheck=false}
\begin{longtable}{@{}
    >{\raggedleft\arraybackslash}p{\dimexpr0.055\linewidth-\tabcolsep\relax}
    >{\raggedright\arraybackslash}p{\dimexpr0.475\linewidth-2\tabcolsep\relax}
    *{2}{>{\centering\arraybackslash}p{\dimexpr0.1567\linewidth-2\tabcolsep\relax}}
    >{\centering\arraybackslash}p{\dimexpr0.1566\linewidth-\tabcolsep\relax}
@{}}
\caption{Task-level recorded completion of Grok with eight independent full-harness attempts per task, without access to records from other attempts. \emph{Single} reproduces the Grok column of Table~\ref{tab:task-completion}; \emph{Passing runs} counts attempts that complete the task under formal evaluation; \emph{Any of 8} marks tasks completed by at least one attempt. A dash denotes no recorded completion, including unfinished evaluation.}
\label{tab:independent-attempts}\\
\toprule
\textbf{ID} & \textbf{Mechanism} & \textbf{Single} & \textbf{Passing runs} & \textbf{Any of 8} \\
\midrule
\endfirsthead
\multicolumn{5}{@{}l}{\textit{\tablename~\thetable{} continued}}\\[4pt]
\toprule
\textbf{ID} & \textbf{Mechanism} & \textbf{Single} & \textbf{Passing runs} & \textbf{Any of 8} \\
\midrule
\endhead
\midrule
\multicolumn{5}{r@{}}{\textit{Continued on next page}}\\
\endfoot
\bottomrule
\endlastfoot
1 & Hinged toolbox & --- & 2/8 & \(\checkmark\) \\
2 & Rack-and-pinion gate & \(\checkmark\) & 7/8 & \(\checkmark\) \\
3 & Four-bar wiper & \(\checkmark\) & 7/8 & \(\checkmark\) \\
4 & Hand-cranked egg beater & \(\checkmark\) & 7/8 & \(\checkmark\) \\
5 & Self-locking utility knife & \(\checkmark\) & 0/8 & --- \\
6 & Folding drying rack & --- & 0/8 & --- \\
7 & Bicycle & --- & 0/8 & --- \\
8 & Pedal-operated bin & \(\checkmark\) & 7/8 & \(\checkmark\) \\
9 & Mechanical alarm clock & --- & 0/8 & --- \\
10 & Kitchen tongs & --- & 2/8 & \(\checkmark\) \\
11 & Retractable ballpoint pen & --- & 0/8 & --- \\
12 & Nail clipper & --- & 3/8 & \(\checkmark\) \\
13 & One-way ratchet screwdriver & --- & 0/8 & --- \\
14 & Linearly actuated scissor lift & --- & 0/8 & --- \\
15 & Equal-arm balance & \(\checkmark\) & 6/8 & \(\checkmark\) \\
16 & Toggle clamp / press & \(\checkmark\) & 6/8 & \(\checkmark\) \\
17 & Dry piston-pump mechanism & \(\checkmark\) & 8/8 & \(\checkmark\) \\
18 & Three-digit combination lock & --- & 4/8 & \(\checkmark\) \\
19 & Two-digit mechanical counter & --- & 0/8 & --- \\
20 & Three-ball dispenser & --- & 0/8 & --- \\
21 & Gravity timer & \(\checkmark\) & 0/8 & --- \\
22 & Slider-crank push rod & \(\checkmark\) & 8/8 & \(\checkmark\) \\
23 & Gravity-pawl turntable & --- & 3/8 & \(\checkmark\) \\
24 & Sliding-bolt gate & \(\checkmark\) & 8/8 & \(\checkmark\) \\
25 & Three-stage gear reducer & --- & 7/8 & \(\checkmark\) \\
26 & Desktop display turntable & \(\checkmark\) & 8/8 & \(\checkmark\) \\
27 & Small swing gate & \(\checkmark\) & 8/8 & \(\checkmark\) \\
28 & Tilting desktop mirror & \(\checkmark\) & 1/8 & \(\checkmark\) \\
29 & Folding wall hook & \(\checkmark\) & 8/8 & \(\checkmark\) \\
30 & Spring clothespin & \(\checkmark\) & 1/8 & \(\checkmark\) \\
31 & Scissors & \(\checkmark\) & 8/8 & \(\checkmark\) \\
32 & Ratchet hand winch & --- & 0/8 & --- \\
33 & Spring-latch door lock & \(\checkmark\) & 1/8 & \(\checkmark\) \\
34 & Reversible dual-pawl ratchet wrench & --- & 0/8 & --- \\
35 & Full-cycle ratchet crimper & --- & 0/8 & --- \\
36 & Geneva indexing table with overload clutch & --- & 0/8 & --- \\
37 & Folding shopping cart & --- & 0/8 & --- \\
38 & Mechanical \(4\times4\) puzzle cube & --- & 7/8 & \(\checkmark\) \\
39 & Twin-cylinder engine linkage & \(\checkmark\) & 2/8 & \(\checkmark\) \\
40 & Four-station indexing table & --- & 0/8 & --- \\
41 & Synchronized parallel gripper & --- & 0/8 & --- \\
42 & Releasable reduction winch & --- & 0/8 & --- \\
43 & Baby stroller & --- & 0/8 & --- \\
44 & Height- and tilt-adjustable worktable & --- & 0/8 & --- \\
45 & Coin sorter and counter & --- & 0/8 & --- \\
46 & Interchangeable-tune music box & --- & 0/8 & --- \\
\midrule
\multicolumn{2}{@{}l}{\textbf{Total completed}} &
\textbf{19/46} & \textbf{129/368} & \textbf{24/46} \\
\end{longtable}
\endgroup

\section{Details on 3D-Printed Mechanical Prototypes}
\label{app:printed-mechanisms}

\begin{figure}[!htbp]
\centering
\begin{minipage}[t]{0.487\linewidth}
\centering
\includegraphics[width=\linewidth,trim=160bp 65bp 88bp 0bp,clip]{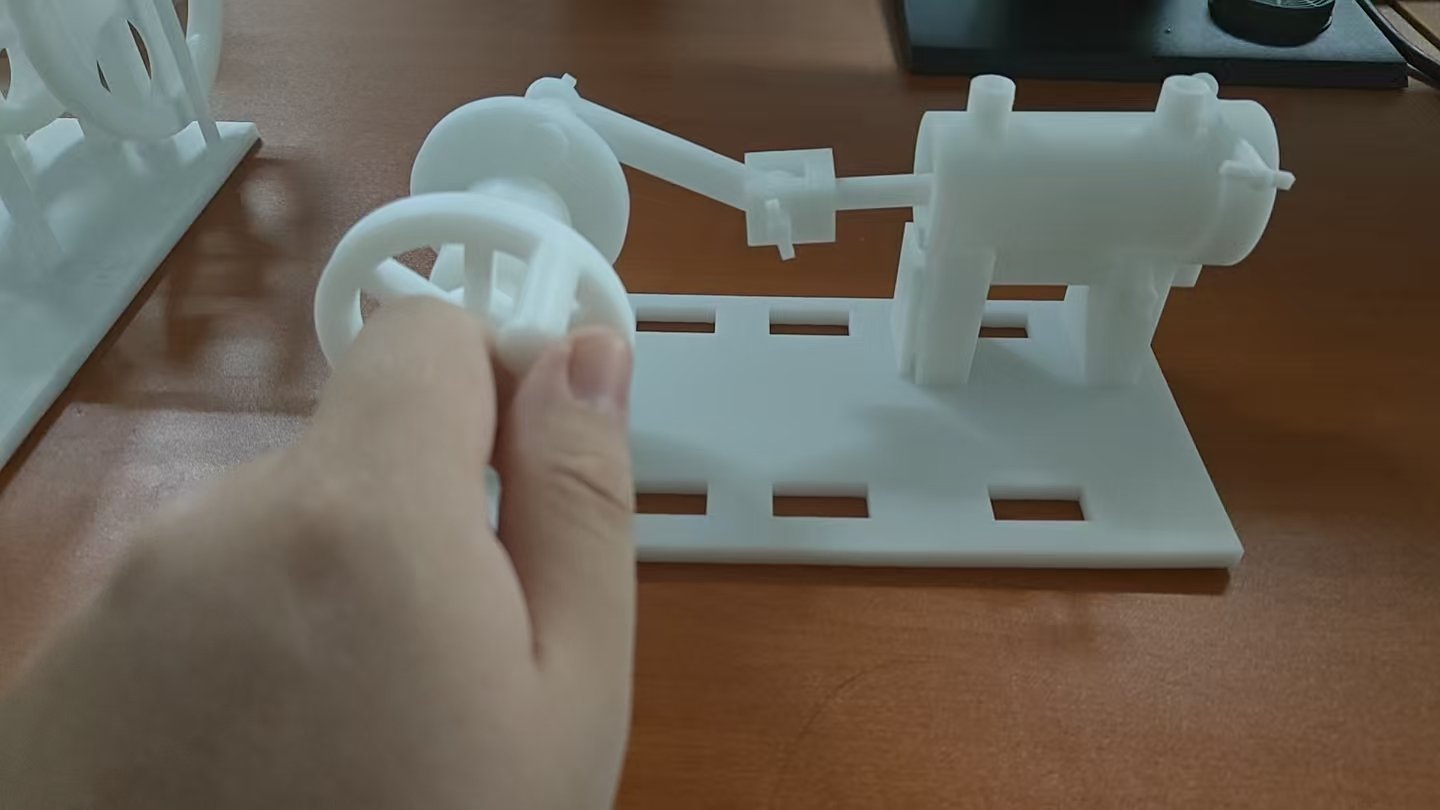}
\par\vspace{3pt}
{\small\textbf{(a) Piston: crank position 1}}
\end{minipage}\hfill
\begin{minipage}[t]{0.487\linewidth}
\centering
\includegraphics[width=\linewidth,trim=120bp 30bp 72bp 0bp,clip]{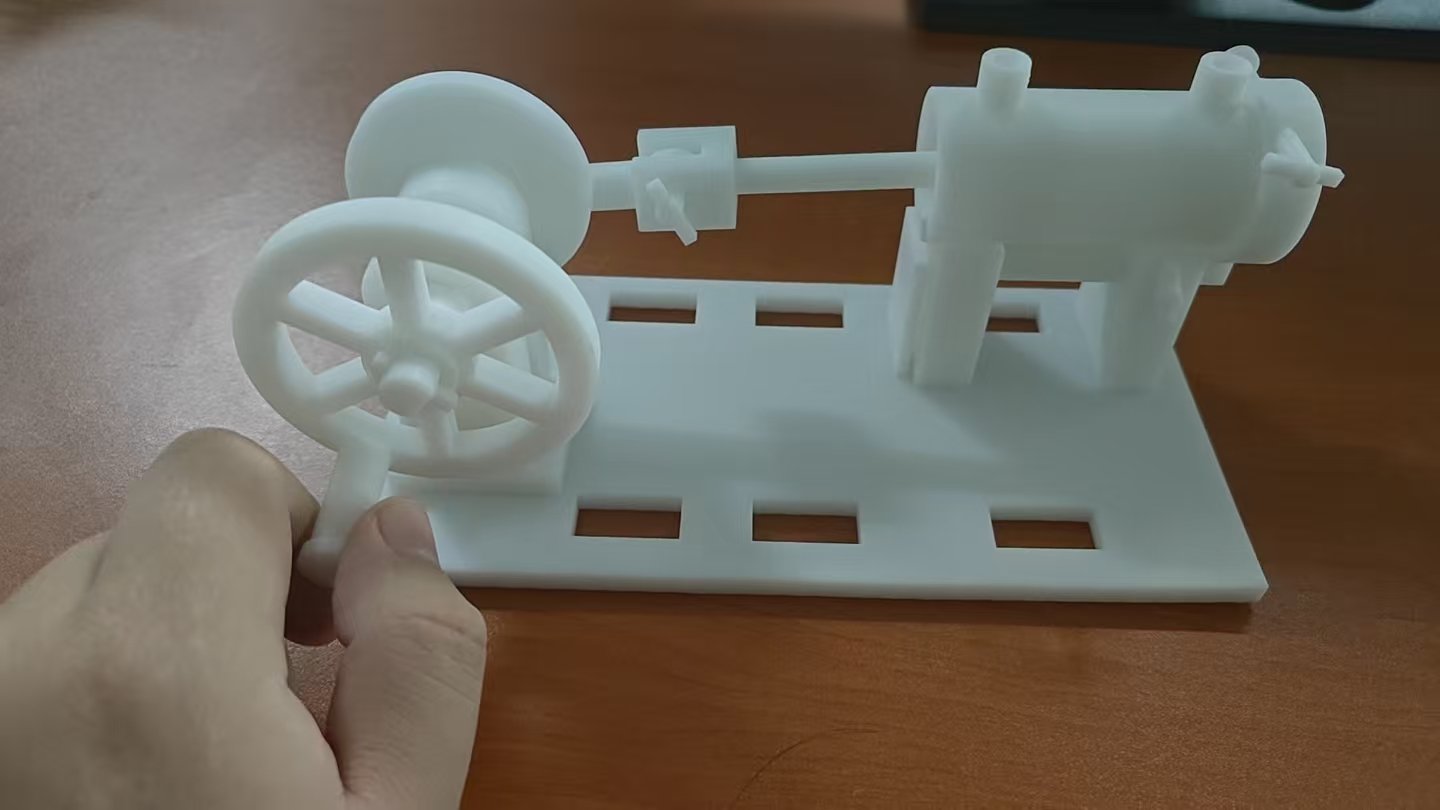}
\par\vspace{3pt}
{\small\textbf{(b) Piston: crank position 2}}
\end{minipage}

\vspace{12pt}
\begin{minipage}[t]{0.487\linewidth}
\centering
\includegraphics[width=\linewidth,trim=180bp 45bp 76bp 25bp,clip]{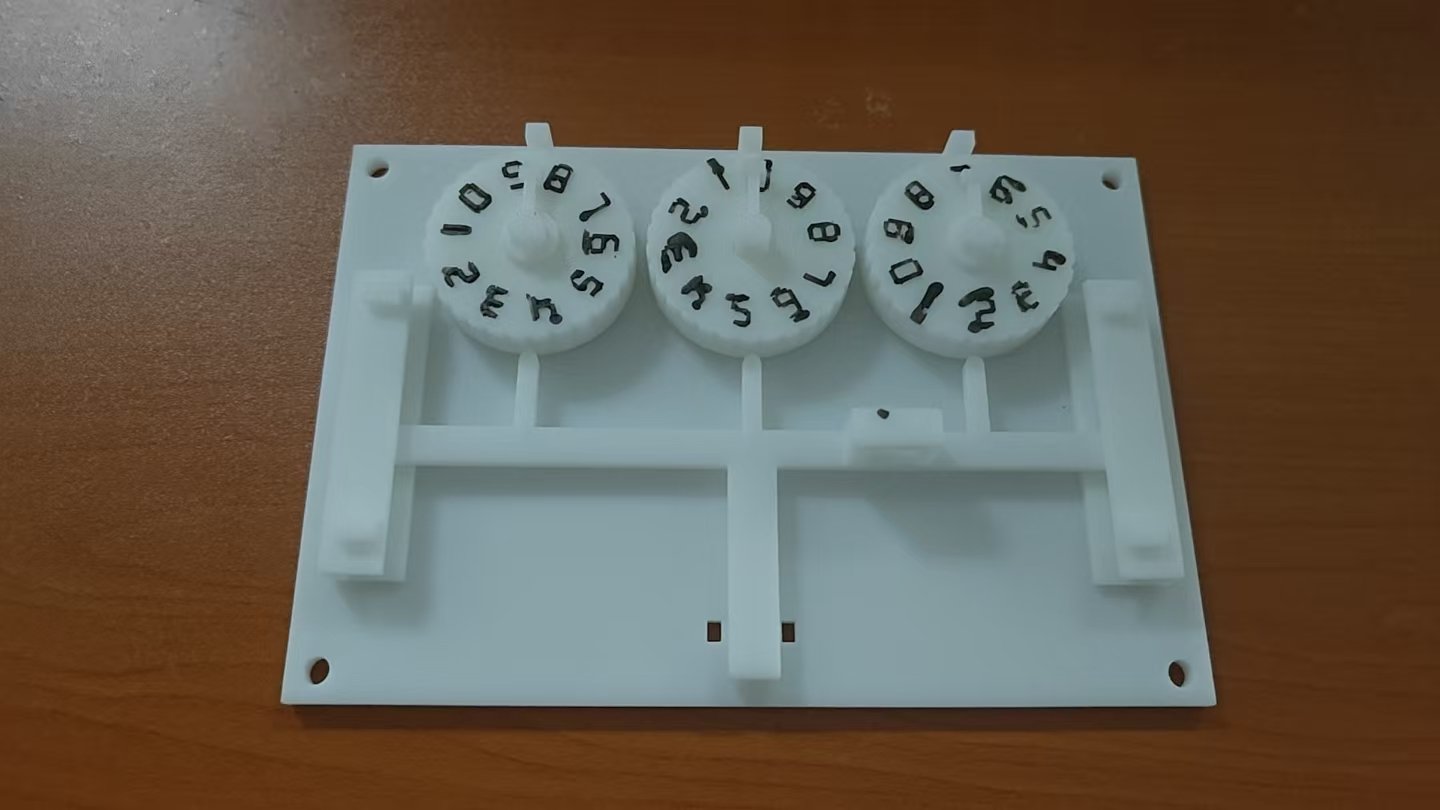}
\par\vspace{3pt}
{\small\textbf{(c) Three-digit combination lock}}
\end{minipage}\hfill
\begin{minipage}[t]{0.487\linewidth}
\centering
\includegraphics[width=\linewidth,trim=120bp 0bp 24bp 0bp,clip]{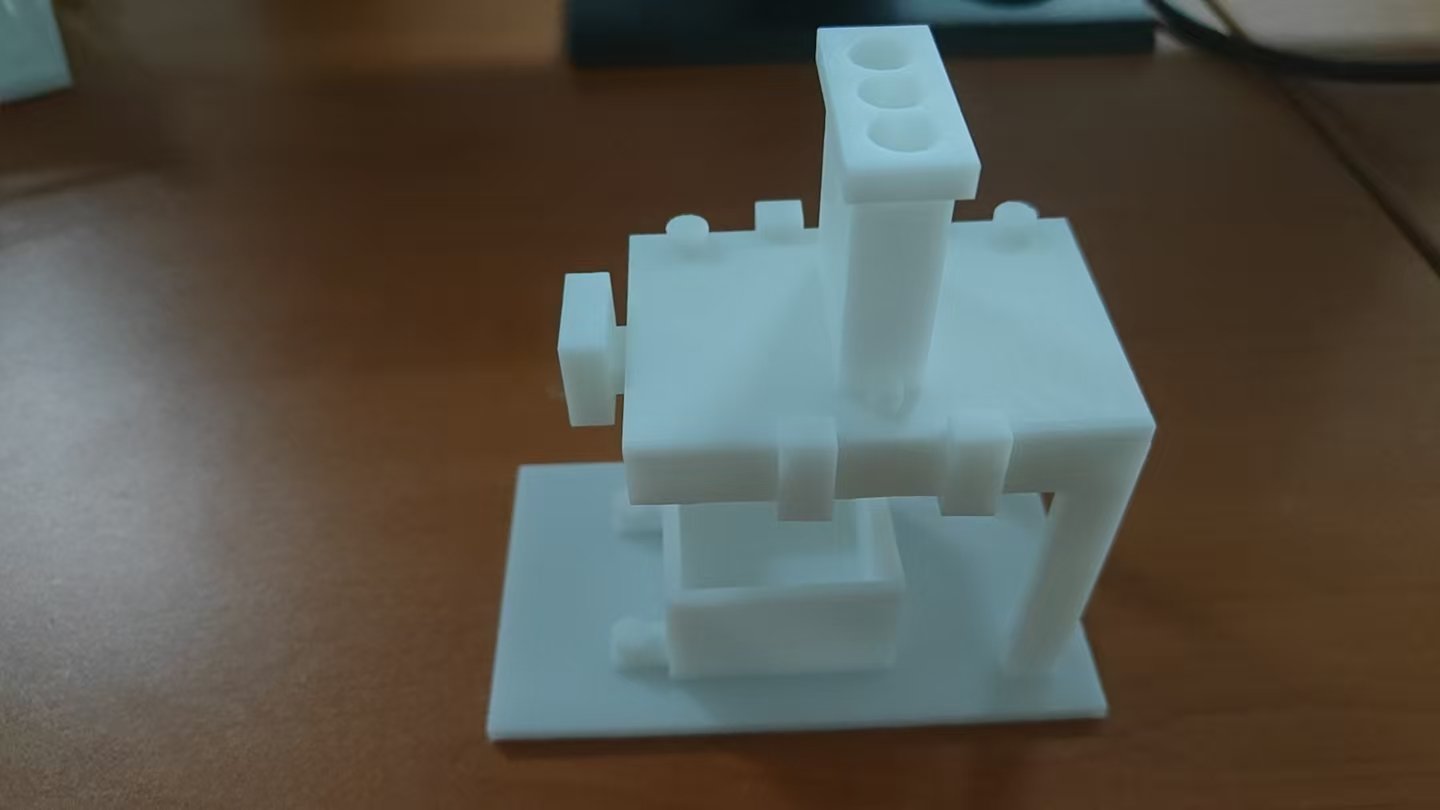}
\par\vspace{3pt}
{\small\textbf{(d) Three-ball dispenser}}
\end{minipage}
\caption{\textbf{Printed prototypes for piston motion, code-controlled
locking, and ball dispensing.} (a--b) Task~17: two crank positions of
the piston linkage. (c) Task~18: three numbered code wheels and the
locking bar. (d) Task~20: the loading channels, dispensing slider,
and receiving tray.}
\label{fig:printed-prototypes-components}
\label{fig:printed-mechanisms}
\end{figure}

\paragraph{Piston (Task~17).}
The crank and connecting rod drive a guided piston back and forth.
The task requires repeated strokes with the specified travel and
return position, including motion against resistance.
Panels~(a--b) show the printed assembly at two crank positions.

\paragraph{Combination lock (Task~18).}
Three independently adjustable wheels encode the combination.
The task requires unlocking at the correct code, re-locking after
reinserting the locking bar and scrambling the wheels, and blocking
withdrawal when any one digit is incorrect.

\paragraph{Ball distributor (Task~20).}
The slider meters balls from the loading channels into the receiving
tray. The task requires three balls per cycle, cumulative outputs of
3, 6, and 9 across three cycles, and retention of the remaining balls
between operations.

\paragraph{Two-digit counter (Task~19).}
The task requires one increment per forward push-rod stroke,
correct carry through $09\rightarrow10$ and $19\rightarrow20$,
and retention of the reading during the return stroke and at rest.
The two views expose the digit wheels and the linkage that advances them.

\paragraph{Gravity timer (Task~21).}
The prototype implements a nominal 10-second gravity timer.
The task requires autonomous operation after release, triggering
within the specified time window, retention of the visible indication,
and repeat operation after reset.

\paragraph{Cup-access machine (open-ended task).}
The design addresses access to a cup stored behind other cups.
Turning the handle moves the paired side linkages and raises the cup
holder into reach. Panel~(c) shows the initial arrangement; the
supplementary video shows operation of the printed linkage and cup
retrieval.

\end{document}